\documentclass[letterpaper]{article} % DO NOT CHANGE THIS
\usepackage{aaai2026}  % DO NOT CHANGE THIS
\usepackage{times}  % DO NOT CHANGE THIS
\usepackage{helvet}  % DO NOT CHANGE THIS
\usepackage{courier}  % DO NOT CHANGE THIS
\usepackage[hyphens]{url}  % DO NOT CHANGE THIS
\usepackage{graphicx} % DO NOT CHANGE THIS
\usepackage{natbib}  % DO NOT CHANGE THIS AND DO NOT ADD ANY OPTIONS TO IT
\usepackage{caption} % DO NOT CHANGE THIS AND DO NOT ADD ANY OPTIONS TO IT
\usepackage{multirow}
\usepackage{booktabs}
\usepackage{enumitem}
\usepackage{array}
\usepackage[linesnumbered,ruled,vlined,noend]{algorithm2e}
\usepackage{mathcommands}

\newtheorem{problem}{Problem}
\newcommand{\dflood}{\textsc{Diff-Sparse}}

\newcommand{\numquad}[1][1]{\hspace*{#1em}\ignorespaces}

\newcommand{\conv}{\textsc{conv}}
\newcommand{\linear}{\textsc{linear}}
\newcommand{\sinusoid}{\textsc{sinusoid}}

\title{Towards High Resolution Probabilistic Coastal Inundation Forecasting from Sparse Observations}

\author {
    Kazi Ashik Islam\textsuperscript{\rm 1,},
    Zakaria Mehrab\textsuperscript{\rm 1},
    Mahantesh Halappanavar\textsuperscript{\rm 2},
    Henning Mortveit\textsuperscript{\rm 1},
    Sridhar Katragadda\textsuperscript{\rm 3},
    Jon Derek Loftis\textsuperscript{\rm 4},
    Stefan Hoops\textsuperscript{\rm 1},
    Madhav Marathe\textsuperscript{\rm 1}
}
\affiliations {
    \textsuperscript{\rm 1}Biocomplexity Institute, University of Virginia\\
    \textsuperscript{\rm 2}Pacific Northwest National Laboratory\\
    \textsuperscript{\rm 3} Department of Communications and Information Technology, City of Virginia Beach\\
    \textsuperscript{\rm 4} Virginia Institute of Marine Science, College of William and Mary\\
    \{ki5hd, zm8bh\}@virginia.edu,
    hala@pnnl.gov, Henning.Mortveit@virginia.edu, SKatraga@vbgov.com, jdloftis@vims.edu, \{shoops, marathe\}@virginia.edu
}
\usepackage{bibentry}
\begin{document}

\maketitle

\begin{abstract}
Coastal flooding poses increasing threats to communities worldwide, necessitating accurate and hyper-local inundation forecasting for effective emergency response. However, real-world deployment of forecasting systems is often constrained by sparse sensor networks, where only a limited subset of locations may have sensors due to budget constraints. 
To approach this challenge, we present \dflood{}, a masked conditional diffusion model designed for probabilistic coastal inundation forecasting from sparse sensor observations. 
\dflood{} primarily utilizes the inundation history of a location and its neighboring locations from a context time window as spatiotemporal context. The fundamental challenge of spatiotemporal prediction based on sparse observations in the context window is addressed by introducing a novel masking strategy during training. 
Digital elevation data and temporal co-variates are utilized as additional spatial and temporal contexts, respectively. A convolutional neural network and a conditional UNet architecture with cross-attention mechanism are employed to capture the spatiotemporal dynamics in the data.  
We trained and tested \dflood{} on coastal inundation data from the Eastern Shore of Virginia and
systematically assessed the performance of \dflood{} across different sparsity levels ($0\%, 50\%, 95\%$ missing observations). Our experiment results show that \dflood{} achieves upto $62\%$ improvement in terms of two forecasting performance metrics 
compared to existing methods, at $95\%$ sparsity level.
Moreover, our ablation studies reveal that digital elevation data becomes more useful at high sparsity levels compared to temporal co-variates.
\end{abstract}

\begin{links}
    \link{Code}{https://github.com/KAI10/Diff-Sparse}
    % \link{Datasets}{https://aaai.org/example/datasets}
    % \link{Extended version}{https://aaai.org/example/extended-version}
\end{links}

\section{Introduction}

% \textbf{Forecasting Coastal Inundation: A Social Good Problem}
The projection in sea level rise have highlighted the increasing vulnerability of many coastal regions to inundation~\citep{wuebbles2017climate}. 
% One recent example is the flash flood incident in Texas Hill county~\cite{cbstexas}, causing more than 100 fatalities. 
Among these regions, the East Coast in US is a region of particular policy concern due to its geography and low-lying topography~\citep{ezer2014accelerated}. In recent periods, this region has been experiencing frequent inundation events ~\citep{ezer2020analysis}, causing disruptions in daily life (e.g., traffic, economic)~\citep{jacobs2018recent}. In future, this situation will only continue to decline due to the combined effect of climate change, sea level rise, and urbanization~\citep{swain2020increased}. Therefore, forecasting coastal inundation is of utmost importance from a policy standpoint for undertaking risk prevention measurement and safe evacuation of residents during emergencies ~\citep{yang2019incorporating}. %\zakaria{Add texas kerrfield flooding event citation, news article?}

The problem of forecasting coastal inundation from sparse observation was brought to our attention through conversations with staff at the City of Virginia Beach. The region faces challenges with coastal flooding which causes slow-down and/or road closures; disrupting the day-to-day activities of the inhabitants. A fast and high-resolution flood forecasting model would enable city officials to identify localized flood risks, resulting in more effective preparedness and response. 
% In this paper, we approach the problem of designing such a model under sparse observation setting. 
% It is a joint-work of a multidisciplinary team, including computer and data scientists, academic experts in hydro-dynamic models, members of coastal resources management centers, and officials at the City of Virginia Beach.
% , and members of established coastal flooding centers.
% \textbf{Why Forecasting Coastal Inundation is important?} The water level of along the east coast is rising. It has already caused damage (statistics on damage). 
In order to effectively identify inundated locations, policymakers deploy sensors at various locations~\cite{tao2024review}. These sensors are responsible for providing periodical inundation reports and other variables of interest in the surrounding area. However, there are several challenges associated with these sensor placements. \emph{First,} due to budget and resource constraints, the number of sensors that can be deployed is limited~\cite{saad2024scalable, karagulian2019review}. \emph{Second,} the readings from the sensor are often associated with noise due to mechanical, environmental or calibration issues~\cite{barrenetxea2008assessing,mydlarz2024floodnet}. \emph{Third,} due to the difficulty associated with finding an optimal location assignment of spatial sensors~\cite{krause2008near} or changes carried out by the stakeholders, the locations of these sensors may change from time to time~\cite{garg2012learning}. As a result, a forecasting method that relies on sensor data but does not account for the sparsity and uncertain nature of sensor placements will not be reliable~\cite{tao2024review}.
% \zakaria{add citations here after careful review

% Constrained coverage for mobile sensor networks -- mobile sensor placement challenges

% https://jmlr.org/papers/volume9/krause08a/krause08a.pdf -- limited budget sensor placement}

% \begin{figure}[t]
%     \centering
%     \includegraphics[width=\columnwidth]{ACM-SIGSPATIAL-2025/Figures/quinby_bridge_inundated.pdf}
%     \caption{Flooding on Quinby Bridge Road, Accomack County, VA on April 3, 2024 (Data source: Tidewatch Maps \citep{TidewatchMaps}.)}
%     \label{fig:road_inundation}
% \end{figure}

% \noindent
% \textbf{Multi-disciplinary Team} \

% \noindent
% \textbf{Technical Challenges}\ 
The traditional method for predicting inundation involves hydro-dynamical, physics-based models (PBM). Commercially available \citep{syme2001tuflow,deltares2014sobek} (e.g., SOBEK, TUFLOW) or open-sourced~\citep{zhang2016seamless} (e.g, SCHISM) hydro-dynamical models can simulate inundation accurately at a high resolution. However, the computation time associated with these PBMs makes them challenging for real-time forecasting~\citep{guo2021urban}. 
Researchers have investigated the use of Deep-Learning (DL) based prediction methods as surrogates to these PBMs~\citep{roy2023application}. However, these models often struggle to: (i) scale in large-scale high-resolution forecasting tasks under sparse availability of data, and (ii) effectively use additional spatiotemporal contexts. Therefore, a surrogate model to PBMs that can make accurate real-time predictions based on sparse sensor observations, will have high relevance to policymakers.  
%\zakaria{or perform poorly in sparse observation setting.}
% nor 
%do they take other contexts into account that may be of importance for modeling physical properties captured in these physics-based models.
% are they trained to learn physical properties implicit in the PBMs.

% \zakaria{Have to rewrite the below part}

% \zakaria{why diffusion model is good in sparse setting? Is it solely the ability to condition on contexts or something else?}

% \noindent
% \textbf{Our Contributions}\ 

Motivated by the above observations, we explore Diffusion Models~\citep{sohl2015deep} for high-resolution coastal inundation forecasting task. The reason is two-fold: their potential to model physical properties~\citep{ahmad2024calobench,amram2023denoising}, and their ability to utilize conditioning contexts of various dimensions~\citep{Rombach_2022_CVPR,zhang2023adding} that can improve prediction quality in the presence of sparse observations.
Although diffusion models were originally developed for image generation, they have been adapted for time-series and spatiotemporal forecasting~\citep{rasul2021autoregressive,wen2023diffstg}. However, the existing methods struggle in terms of computational scalability and performance, under sparse availability of data. %We address these challenges by presenting a scalable probabilistic spatiotemporal forecasting method, and use coastal inundation of the Eastern Shore as case study for its application.
Using the coastal inundation of the Virginia Eastern Shore as case study, we address these challenges by making the following contributions:
% a scalable method that can take both spatial and temporal context into account. At the same time it can incorporate domain-specific constraints/objectives into its training objective. We use coastal inundation forecasting as a motivating example for its application.
% leverages spatial contexts to create low-dimensional latent representations and novel domain-specific constraints to train the diffusion model for spatiotemporal forecasting, using coastal inundation as a motivating example. 
%Specifically, we make the following contributions in this paper.

\begin{itemize}[leftmargin=*,itemsep=0pt]
    \item We formulate the problem of probabilistic inundation forecasting from sparse observations as a conditional distribution function learning problem and present \dflood{}, a scalable model to learn this function. \dflood{} is trained using data from physics-based hydro-dynamic model; however, during inference it makes probabilistic forecasts based on sparse sensor observations and additional spatiotemporal contexts (e.g., elevation, temporal co-variates). % This allows \dflood{} to make hyper-local forecasts in real-time.

    \item We introduce a novel masking strategy to train \dflood{} which enables it to make forecasts from sparse observations. Our robust training strategy enables \dflood{} to make accurate spatiotemporal forecasts from different spatial placement configurations of the sensors, without requiring training the model from scratch. 
    
    \item Extensive evaluation of our method against several baseline methods under sparse observation underscores the superiority of our method at large-scale high resolution inundation forecasting tasks. 
    % and its aspect of being used as a decision-making tool for policymakers. 
    Specifically, we observe upto $62\%$ improvement in prediction performance over existing methods in terms of two predictive performance metrics. 
    
    \item We perform extensive ablation studies to highlight the utility of different context data (e.g., sensor data, elevation, temporal co-variates) at varying sparsity levels. 
    % and the importance of different conditional contexts in improvement of the learning capability.
    
    % \item Furthermore, we demonstrate the scalability of our model compared to other baselines in large settings, underscoring its utility in real-time forecasting in large-scale scenarios.
    %\item A key novelty of this work lies in including domain-specific knowledge to accelerate the model performance. Specifically, we guide the model with knowledge from three different perspectives during training. \emph{First,} we use patch embeddings to guide the model on region representation. \emph{Second,} we guide the model in capturing the spatial correlation between inundation and elevation, a key physical property within the problem domain, by extracting spatial feature embedding. \emph{Finally,} we introduce a novel loss in the objective function, termed \emph{Elevation Consistency Loss} (\ecl{}), to capture the physical constraints in the data distribution. 
\end{itemize}

\iffalse
It is worth highlighting that our model can be used to answer inundation scenario-based queries relevant to policymakers. For example, queries of the form: \textit{``What is the probability that a particular region will get inundated by 1 feet within the next 12 hours?''}, can be answered by using our model through probabilistic forecasting (see section~\ref{sec:discussion} for details). 
\fi

\medskip 

\section{Related Work}

We present an abridged version of the most relevant literature in this section. The supplementary material of this paper covers the literature more extensively.

\noindent
\textbf{Physcis-based Models:} 
%Traditionally, various Physics Based hydrodynamic Models (PBM) have been employed to simulate inundation. SOBEK~\citep{deltares2014sobek} is a 2D model that solves a Saint-Venant flow equation and shallow water equation to determine the water level of rectangular grids. TELEMAC-2D~\citep{vu2015two} uses a similar method but is more suitable for coastal areas. Polymorphic models like SCHISM~\citep{zhang2016seamless} can capture water flow at very high resolution. However, the computation time associated with solving these complex equations at high spatial resolution makes these PBMs infeasible for real-time forecasting.
Traditionally, various Physics-Based hydrodynamic Models (PBMs) have been employed to simulate inundation. SOBEK~\cite{deltares2014sobek} is a two-dimensional model that solves the Saint-Venant flow equations and shallow water equations to determine water levels across rectangular grids. Similarly, TELEMAC-2D~\cite{vu2015two} employs comparable method specifically tailored for coastal applications. Polymorphic models such as SCHISM~\cite{zhang2016seamless} captures water flow at very high resolutions. However, the substantial computational time associated with solving these complex equations at high spatial resolution renders these PBMs impractical for real-time forecasting applications.

\noindent
\textbf{Deep Learning}
% DL methods are a popular choice in surrogating hydrodynamic models. Regression-based methods~\citep{zahura2022predicting}, recurrent neural networks~\citep{roy2023application} and hybrid ML-based methods~\citep{zanchetta2022hybrid} are some of the methods that have been attempted as possible surrogates. However, these models cannot inherently take spatial context into account. The DCRNN method proposed by~\citep{li2017diffusion} for traffic forecasting is the state-of-the-art DL method for spatiotemporal point forecasting, where bidirectional random walk is used to capture spatial dependency and Seq2Seq architecture is used to capture temporal dependency. BayesNF~\citep{saad2024scalable} is the state-of-the-art DL spatiotemporal forecasting model, which is built using a Bayesian Neural Network. By assigning a prior distribution to the model parameters and adjusting the posterior based on observation, BayesNF essentially learns to map multivariate spatiotemporal points to a continuous real-valued field. 
Researchers have explored Deep Learning (DL) methods for developing surrogates to PBMs. Regression-based methods~\cite{zahura2022predicting}, recurrent neural networks~\cite{roy2023application}, and hybrid machine learning-based methods~\cite{zanchetta2022hybrid} have been emerged as popular choices, among others. However, these models inherently struggle to incorporate spatiotemporal contexts effectively. Originally developed for traffic forecasting, the Diffusion Convolutional Recurrent Neural Network (DCRNN) method proposed by Li et al.~\citeyear{li2017diffusion} captures some aspect of spatiotemporal contexts by utilizing bidirectional random walks and Sequence-to-Sequence architecture. More recently, BayesNF~\cite{saad2024scalable} has emerged as the state-of-the-art DL spatiotemporal forecasting model, constructed using a Bayesian Neural Network framework. By assigning prior distributions to model parameters and adjusting the posterior based on observations, BayesNF effectively learns to map multivariate spatiotemporal points to continuous real-valued fields.

\noindent
\textbf{Diffusion Model} The field of generative modeling has been dominated by the diffusion model for quite some time. Although it had primarily been studied for image synthesis~\citep{ho2020denoising,dhariwal2021diffusion,austin2021structured}, its fascinating ability to use flexible conditioning to guide the generation process has been utilized in other domains like video generation~\citep{yang2023diffusion}, prompt-based image generation~\citep{ramesh2022hierarchical}, text-to-speech generation~\citep{yang2023diffsound}. TimeGrad~\citep{rasul2021autoregressive} was the pioneering method that approached the time-series forecasting problem by utilizing a diffusion model. Other models like CSDI~\citep{tashiro2021csdi}, DPSD-CSPD~\citep{bilovs2023modeling}, TDSTF~\citep{Chang2024} were later introduced. These models try to predict the noise added during the forward process of the diffusion model ($\epsilon$-parameterization). An alternative parameterization is to predict the ground-truth data ($x$-parameterization). 
% We utilize the $x$-parameterization to accommodate domain-specific constraints in the training objective. 
Different from models mentioned above, $D^3$VAE~\citep{li2022generative} uses coupled diffusion process and bidirectional variational autoencoder to better forecast in longer horizon and improve interpretability. The State-of-the-Art Diffusion-based model DiffSTG~\citep{wen2023diffstg} devised a modified architecture for the reverse process of diffusion model combining UNet and GNN to capture temporal and spatial dependencies, respectively. However, both methods scale poorly with the growth of variables.

% \zakaria{
% \begin{itemize}
%     \item hydrological models -- done
%     \item regression-based models --done
%     \item DL papers -- goodall --done
%     \item Diffusion model -- classical
%     \item Diffusion -- other applications. (time series forecasting)
%     \item forecasting-based policy paper
% \end{itemize}}
%\zakaria{\cite{tashiro2021csdi,bilovs2023modeling,Chang2024} cite and mention them using theta parameterization and sparse time series. Mention that we need x parameterization for incorporating domain-sepcific constraint. Shorten hydrology section} 
\section{Problem Formulation}
% (MVM): Define the problem precisely here
Formally, we define the inundation forecasting from sparse observations problem as follows.

\begin{problem}
\label{prob:multipatch_forecast}
\textbf{Multi-patch Probabilistic Inundation Forecasting.} Let $\cP = \{P_1, P_2, ..., P_K\}$ denote a set of two-dimensional rectangular grids each of size $D \times D$. We refer to $P_k$ as the $k^{th}$ `patch'.
Let $\cM = \{\Mb_1, \Mb_2, ..., \Mb_K\}$ denote a set of binary matrices, henceforth referred to as `sensor masks'. If $\Mb_{k}(i, j) = 1$ then the cell $(i, j)$ of patch $P_k$ has a water-level sensor; $\Mb_{k}(i, j) = 0$ means cell $(i, j)$ of patch $P_k$ does not have a water-level sensor.
Let ${}_k\yb_t^0 \in \RR^{D \times D}$ denote the inundation matrix of patch $P_k$ at timestep $t$. If $\Mb_{k}(i,j) = 1$ then ${}_k\yb_{ij,t}^0 \in \RR$ contains the inundation level on cell $(i, j)$ of patch $P_k$ at time step $t$; $i, j \in \left\{1, ..., D\right\}, t \in \{1, ..., T\}$. However, for cells $(i, j)$ where $\Mb_{k}(i, j) = 0$, ${}_k\yb_{ij,t}^0, t \in \{1, ..., T\}$ has missing value.
Let ${}_k\xb_t^0 \in \RR^{D \times D}$ denote the complete inundation matrix of patch $P_k$ at timestep $t$, where there are no missing values.
Let $\sbb_k \in \RR^{D \times D}$ and ${}_k\zb_t \in \RR^{l}, t \in \{1, \ldots, T\}$ denote the elevation matrix and time series of co-variates associated with patch $P_k$, respectively.
Let $t_0:t_{c+l}$ denote the increasing sequence of timesteps ${t_0, t_1, ..., t_{c+l}}$, where $c, l > 0$.
We aim to learn the conditional distribution function $\textstyle q({}_k\xb_{t_c:t_{c+l}}^0 | {}_k\yb_{t_0:t_{c-1}}^0, {}_k\zb_t, \sbb_k, \Mb_k)$.
\end{problem}

% \begin{problem}
% \label{prob:multipatch_forecast}
% \textbf{Multi-patch Probabilistic Inundation Forecasting.} Let $\cP = \{P_1, P_2, ..., P_k\}$ denote a set of two-dimensional patches of size ${D \times D}$.
% Let $\cM = \{\Mb_1, \Mb_2, ..., \Mb_k\}$ denote the corresponding sensor masks.
% Let ${}_k\yb_t^0 \in \RR^{D \times D}$ denote the inundation matrix of patch $P_k$ at timestep $t \in \{1, \ldots, T\}$, with missing values where $\Mb_{k}(i, j) = 0$. Let ${}_k\xb_t^0 \in \RR^{D \times D}$ denote the complete inundation matrix of patch $P_k$ at timestep $t$, where there are no missing values.
% Let $\sbb_k \in \RR^{D \times D}$ and ${}_k\zb_t \in \RR^{l}, t \in \{1, \ldots, T\}$ denote the elevation matrix and time series of co-variates associated with patch $P_k$, respectively.
% Let $t_0:t_{c+l}$ denote the increasing sequence of timesteps ${t_0, t_1, ..., t_{c+l}}$, where $c, l > 0$.
% We aim to learn the conditional distribution function $\textstyle q({}_k\xb_{t_c:t_{c+l}}^0 | {}_k\yb_{t_0:t_{c-1}}^0, {}_k\zb_t, \sbb_k, \Mb_k)$.
% \end{problem}

Here, we aim to forecast the inundation on all cells in a patch (i.e., $\xb$ ) even though in the context inundation history (i.e., $\yb$) we only know the inundation levels on cells where sensors are placed. Problem \ref{prob:multipatch_forecast} is a probabilistic forecasting problem, as we want to learn the distribution of inundation values instead of making a single prediction. 
% Problem \ref{prob:multipatch_forecast} represents a generalized version of Problem \ref{prob:single_patch_forecast}. Here, we aim to learn a single model that can be utilized to forecast inundation levels on multiple patches/areas. 

The following section presents the basics of a denoising diffusion model used by our method \dflood{} for learning our target distributions. Subsequently, we present details on how we have designed \dflood{} to capture spatial and temporal dynamics of inundation values.
\section{Denoising Diffusion Model}
\label{sec:diffusion_basic}
In this section, we provide a brief overview of denoising diffusion models. A more comprehensive view is provided in the supplementary materials.

Let $\xb^0 \sim q_\chi(\xb^0)$ denote the multivariate training vector from some input space $\chi = \mathbb{R}^D$, where $q_\chi(\xb^0)$ denotes the true distribution. In diffusion probabilistic models, we aim to approximate $q_\chi(\xb^0)$ by a probability density function $p_{\theta}(\xb^0)$ where $\theta$ denotes the set of trainable parameters. Diffusion probabilistic models \citep{sohl2015deep,luo2022understanding} are a special class of Hierarchical Variational Autoencoders \citep{kingma2016improved,sonderby2016ladder} with the form $p_{\theta}(\xb^0) = \int p_\theta(\xb^{0:N})d\xb^{1:N}$; here $\xb^1, \xb^2, ..., \xb^N$ are latent variables. Three key properties of diffusion models are: $(i)$ all of the latent variables  are assumed to have the dimension $D$ as $\xb^0$, $(ii)$ the approximate posterior 
% $q(\xb^{1:N}|\xb^0) = \prod_{n=1}^N q(\xb^n | \xb^{n-1})$
, $$\textstyle q(\xb^{1:N}|\xb^0) = \prod_{n=1}^N q(\xb^n | \xb^{n-1})$$ 
is fixed to a Markov chain (often referred to as the \textit{forward} process), and $(iii)$ the structure of the latent encoder at each hierarchical step is pre-defined as a linear Gaussian model: $$q(\xb^n|\xb^{n-1}) = \cN(\xb^n; \sqrt{1-\beta_n}\xb^{n-1}, \beta_n\Ib).$$ Additionally, the Gaussian parameters ($\beta_1, \beta_2, ..., \beta_n)$ are chosen in such a way that we have: $q(\xb^N) = \cN(\xb^N; \zerob,\Ib)$. 

The joint distribution $p_{\theta}(\xb^{0:N})$ is called the reverse process. It is defined as a Markov chain with learned Gaussian transitions beginning at $p(\xb^N)$: $$\textstyle p_\theta(\xb^{0:N}) = p(\xb^N) \prod_{n=1}^N p_{\theta}(\xb^{n-1}|\xb^n)$$
Each subsequent transition is parameterized as follows:  
\smallskip
\begin{align}
    p_\theta(\xb^{n-1}|\xb^n) = \cN(\xb^{n-1}; \mu_\theta(\xb^n, n), \Sigma_\theta(\xb^n, n)\Ib) \label{eq:backward_gaussian}   
\end{align}
Both $\mu_\theta: \RR^D \times \NN \rightarrow \RR^D$ and $\Sigma_\theta: \RR^D \times \NN \rightarrow \RR^+$ take two inputs: $\xb^n \in \RR^D$ and the noise step $n \in \NN$. The parameters $\theta$ are learned by fitting the model to the data distribution $q_\chi(\xb^0)$. This is done by minimizing the negative log-likelihood via a variational bound:
% \smallskip
\begin{align}
    -\log p_\theta(\xb^0) \leq \EE_{q(\xb^{1:N}|\xb^0)} \log\frac{q(\xb^{1:N}|\xb^0)}{p_\theta(\xb^{0:N})} \label{eq:variational_bound}
\end{align}
% \smallskip
% Here, we get (\ref{eq:jenesen_apply}) by applying Jensen's inequality. 
We indirectly minimize the negative log-likelihood of the data by minimizing the right hand side of (\ref{eq:variational_bound}).

\cite{ho2020denoising} showed that the forward process has the following property: we can sample $\xb^n$ using $\xb^0$ at any noise step $n$ in closed form . Let, $\alpha_n = 1 - \beta_n$, and $\bar{\alpha}_n = \prod_{i=1}^n \alpha_i$. Then, we have:
\begin{equation}
    q(\xb^n | \xb^0) = \cN\left(\xb^n; \sqrt{\bar{\alpha}_n}\xb^0, (1 - \bar{\alpha}_n)\Ib\right). \label{eq:forward_property}
\end{equation}
Based on this property, \cite{ho2020denoising} showed that one possible parameterization of (\ref{eq:backward_gaussian}) is as follows:
\begin{align}
    \mu_\theta(\xb^n, n) &= \frac{\sqrt{\alpha_n}(1 - \bar{\alpha}_{n-1})}{1 - \bar{\alpha_n}} \xb^n + \frac{\sqrt{\bar{\alpha}_{n-1}}\ \beta_n}{1 - \bar{\alpha_n}} \hat{\xb}_\theta(\xb^n, n) \label{eq:x_parameterization}\\
    \Sigma_\theta(\xb^n, n) &= \frac{1 - \bar{\alpha}_{n-1}}{1 - \bar{\alpha_n}} \beta_n \label{eq:x_parameterization_sigma}
\end{align}

Here, $\hat{\xb}_\theta(\xb^n, n)$ is a neural network that predicts $\xb^0$ from noisy vector $\xb^n$ and noise step $n$. \cite{ho2020denoising} then showed that, the right hand side of (\ref{eq:variational_bound}) can be minimized by minimizing:
\begin{align}
    &\EE_{n, q(\xb^n|\xb^0)}\ \frac{1}{2\tilde{\beta}_n}\frac{\bar{\alpha}_{n-1}\ \beta_n^2}{(1 - \bar{\alpha}_n)^2}|| \xb^0 - \hat{\xb}_\theta(\xb^n, n) ||_2^2
\end{align}

Therefore, optimizing a diffusion model can be done  by training a neural network to predict the original ground truth vector from an arbitrarily noisy version of it. 
% To approximately minimize the summation term in the objective (\ref{eq:obj_2}) across all noise steps $n$, we can minimize the expectation over all noise steps as follows:
% \begin{align}
%     &\argmin_\theta \ \EE_{n \sim U[2, N]}\  \EE_{q(\xb^n|\xb^0)}\nonumber\\
%     &\quad\quad\frac{1}{2\tilde{\beta}_n}\frac{\bar{\alpha}_{n-1}\beta_n^2}{(1 - \bar{\alpha}_n)^2}|| \xb^0 - \hat{\xb}_\theta(\xb^n, n) ||_2^2
%     % &\quad\quad\frac{1}{2\tilde{\beta}_n}\frac{\bar{\alpha}_{n-1}\ (1-\alpha_n)^2}{(1 - \bar{\alpha}_n)^2}|| \xb^0 - \hat{\xb}_\theta(\xb^n, n) ||_2^2
%     % &\quad\quad D_{KL}( q(\xb^{n-1}|\xb^n, \xb^0)\ || \ p_\theta(\xb^{n-1}|\xb^n))
% \end{align}
% This can be done by using stochastic samples over noise steps. 
Once trained, to sample from the reverse process $\xb^{n-1} \sim p_\theta(\xb^{n-1}|\xb^n)$, we first sample $\xb^N$ from $\cN(\zerob, \Ib)$. Then, we compute:
\begin{align}
    \xb^{n-1} = \mu_\theta(\xb^n, n) + \sqrt{\Sigma_\theta}\ \ub \label{eq:backward_sample}
\end{align}
Here, $\ub \sim \cN(\zerob, \Ib)$ for $n \in [2, N]$, and $\ub=\zerob$ when $n=1$. $\mu_\theta(\xb^n, n)$, $\Sigma_\theta$ are computed using (\ref{eq:x_parameterization}) and (\ref{eq:x_parameterization_sigma}), respectively. 
% $\xb^0$ is taken as the generated sample.

\section{\dflood{} Method}
\dflood{} is designed for probabilistic inundation forecasting from sparse observations. Following our notations in Problem \ref{prob:multipatch_forecast}, let ${}_k\yb_t^0 \in \RR^{D \times D}$ denote the `sparse' inundation matrix of the patch $P_k$ at time step $t$, with sensor mask $\Mb_k$. Let ${}_k\xb_t^0 \in \RR^{D \times D}$ denote the complete inundation matrix of patch $P_k$ at timestep $t$.
% Throughout the paper, we will refer to ${}_k\xb_t^0$ as the \textit{grid / patch} at timestep $t$, and ${}_k\xb_{ij, t}^0$ as the \textit{grid / patch cell} $(i, j)$ at timestep $t$. 
Towards our goal of learning the conditional distribution $q_\chi({}_k\xb_{t_c:t_{c+l}}^0|{}_k\yb_{t_0:t_{c-1}}^0, {}_k\zb_{t_0:t_{c-1}}, \sbb_k, \Mb_k)$, where $\sbb_k \in \RR^{D \times D}$ is the elevation matrix of patch $P_k$, and ${}_k\zb_t$ denotes a time-series of co-variates associated with $P_k$, we assume:
\begin{align}
    &q_\chi({}_k\xb_{t_c:t_{c+l}}^0|{}_k\yb_{t_0:t_{c-1}}^0, {}_k\zb_{t_0:t_{c-1}}, \sbb_k, \Mb_k) = \nonumber\\ 
    &\quad\quad\quad \textstyle \prod_{t=t_c}^{t_{c+l}} q_\chi({}_k\xb_t^0|{}_k\yb_{t_0:t-1}^0, {}_k\zb_{t_0:t-1}, \sbb_k, \Mb_k) \label{eq:step_by_step_pred}
\end{align}
In \dflood{}, we use a denoising diffusion model to learn the conditional distributions on the right hand side of (\ref{eq:step_by_step_pred}). We refer to ${}_k\yb_{t_0:t-1}^0$, ${}_k\zb_{t_0:t-1}$, $\sbb_k$, and $M_k$ together as \textit{context} data.
\begin{figure}[!b]
    \centering
    \includegraphics[width=\columnwidth]{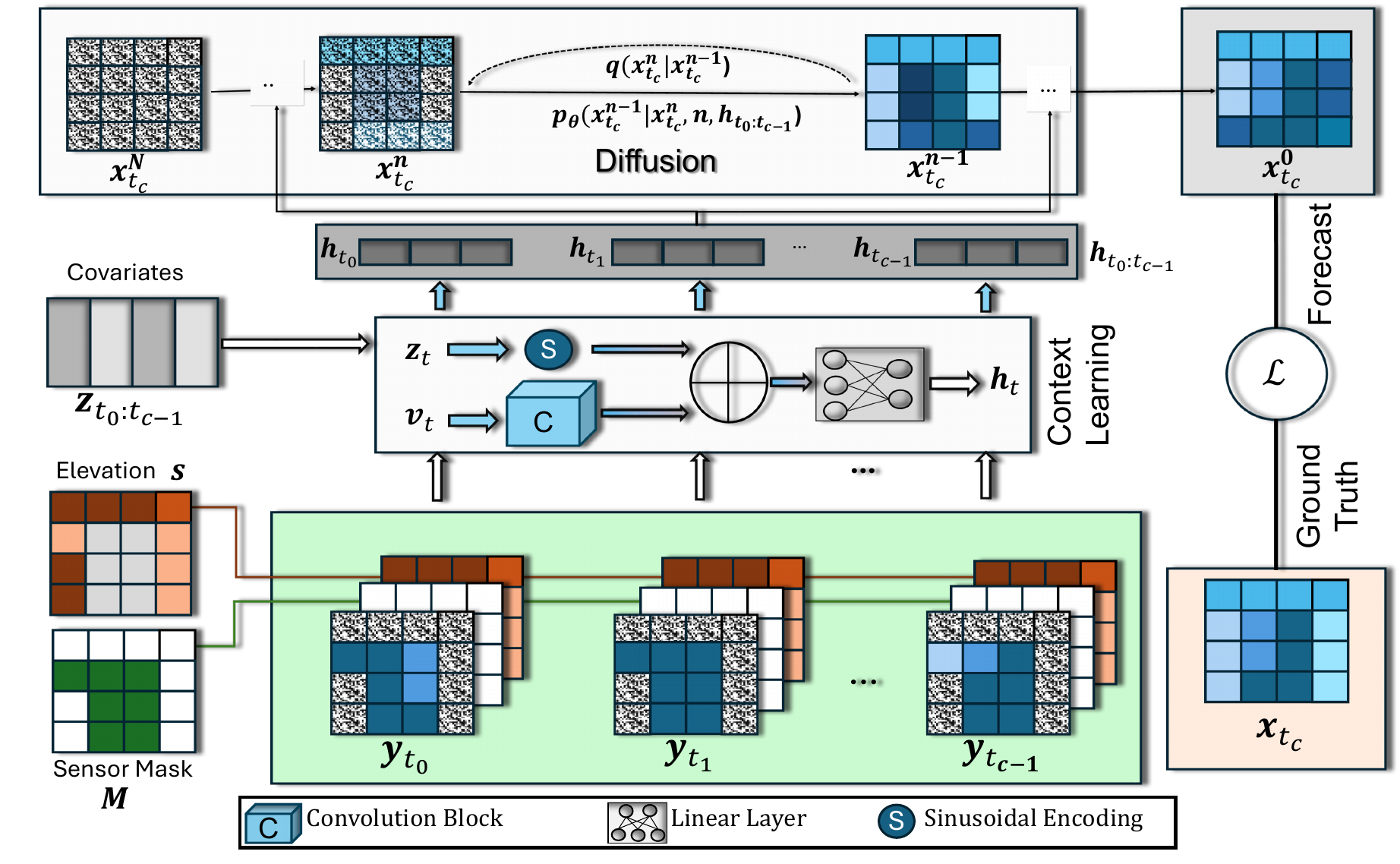}
    \caption{\dflood{} architecture. It has two main components. The first component involves context learning, where the sparse context inundation data ($\yb_{t_0},\yb_{t_1},...,\yb_{t_{c-1}}$), sensor mask ($\Mb$), elevation data ($\sbb$) and temporal covariates ($\zb_{t_0:t_{c-1}}$) are used to generate context embedding $\hb_{t_0:t_{c-1}}$. The second component involves a diffusion model where the corresponding forecast is sampled by conditioning on the context embedding. $q(\xb_{t_c}^n|\xb_{t_c}^{n-1})$ denotes the forward diffusion process; $p_{\theta}(\xb_{t_c}^{n-1}|\xb_{t_c}^n,n,\hb_{t_0:t_{c-1}})$ denotes the reverse diffusion process, parameterized by $\theta$.}
    \label{fig:diff_flood_architecture}
    % \vspace{-0.25cm}
\end{figure}
To capture spatial correlation, we augment the sparse inundation data in the context (i.e., ${}_k\yb^0_{t_0:t-1}$) by concatenating the elevation matrix $\sbb_k$ and the sensor mask $\Mb_k$ to it as additional channels (\ref{eq:append_elev}). The goal is to capture the correlation of elevation and inundation values. We then use convolution blocks to extract spatial features ${}_k\fb_{t_0:t-1}$ (\ref{eq:spatial_embedding}). 
% ${}_k\hb^0_{t_0:t-1}$. We refer to this network as the Spatial Feature Extractor Network (\sfen{}).
\begin{align}
    &{}_k\vb_{t_0 : t-1} = {}_k\yb^0_{t_0 : t-1} \oplus \sbb_k \oplus M_k \label{eq:append_elev}\\
    &{}_k\fb_{t_0:t-1} = \conv{}_{\theta_1}({}_k\vb_{t_0:t-1}) \label{eq:spatial_embedding}
\end{align}
Since coastal inundation is expected to be correlated with tide cycles, we use hour of day, and day of month as temporal features $\zb_t$. We use sinusoidal encoding~\cite{vaswani2017attention} to capture the cyclic nature of these temporal features.
\begin{align}
    {}_k\gb_{t_0:t-1} = \sinusoid{}({}_k\zb_{t_0:t-1})
\end{align}
We concatenate the spatial and temporal features, and then pass it through a linear fully connected layer to get an embedding of the context.
\begin{align}
    {}_k\hb_{t_0:t-1} = \linear{}_{\theta_2}({}_k\fb_{t_0:t-1} \oplus {}_k\gb_{t_0:t-1}) \label{eq:context_embedding}
\end{align}
% Here, $\theta_1, \theta_2$ denote the trainable parameters of the convolution blocks and fully connected layers. 
An overview of \dflood{} architecture is provided in Figure \ref{fig:diff_flood_architecture}. We approximate (\ref{eq:step_by_step_pred}) by the model in (\ref{eq:approx_model}). Here, $\theta$ comprises of the trainable parameters of the convolution blocks ($\theta_1$), fully connected layer ($\theta_2$), and the denoising diffusion model ($\theta_3$).
\begin{align}
    \textstyle \prod_{t=t_c}^{t_{c+l}} p_\theta({}_k\xb_t^0 | {}_k\hb_{t_0:t-1}) \label{eq:approx_model}
\end{align}

As described in the previous section, within the denoising diffusion model, we need to train a neural network $\hat{\xb}_{\theta_3}({}_k\xb^n_t, n, {}_k\hb_{t_0:t-1})$ that predicts ${}_k\xb^0_t$ from noisy vector ${}_k\xb^n_t$, noise step $n$, and context embedding ${}_k\hb_{t_0:t-1}$. We use a conditional UNet for this purpose. 
Within the UNet architecture, we employ a cross-attention conditioning mechanism \citep{Rombach_2022_CVPR} to map the context embedding to the intermediate layers of the UNet. It allows us to predict ${}_k\xb^0_t$ while paying attention to the context embedding.

\subsection{Training}
\begin{algorithm}[!b]
    \caption{Training step on a data point of $P_k$.}
    \label{alg:training}
    \KwIn{
        Data: ${}_k\xb^0_t \sim q({}_k\xb^0_t)$, Context: ${}_k{\xb^0_{t_0:t-1}}$, ${}_k\zb_{t_0:t-1}$, $\sbb_k$.
    }
    \While{not converged}{
        \tcp{Masking}
        Randomly generate a binary sensor mask $\Mb_k$.\\
        \For{$\tau \in [t_0, t-1]$}{ \label{algline:data_masking_start}
            $\ub \sim \cN(\zerob, \Ib)$\\
            ${}_k\yb^0_\tau = {}_k\xb^0_\tau \odot \Mb_k + (\oneb - \Mb_k) \odot \ub$ \label{algline:data_masking_end}
        }
        \tcp{Compute context embedding}
        ${}_k\fb_{t_0:t-1} \gets \conv{}_{\theta_1}({}_k\yb^0_{t_0:t-1} \oplus \sbb_k \oplus \Mb_k)$\\ \label{algline:context_embedding_start}
        ${}_k\gb_{t_0:t-1} \gets \sinusoid{}({}_k\zb_{t_0:t-1})$\\
        ${}_k\hb_{t_0:t-1} \gets \linear{}_{\theta_2}({}_k\fb_{t_0:t-1} \oplus {}_k\gb_{t_0:t-1})$\\ \label{algline:context_embedding_end}
        \tcp{Forward diffusion}
        Initialize $n \sim \text{Uniform}(1, N)$.\\
        ${}_k\xb^n_t \gets \cN\left(\sqrt{\bar{\alpha}_n}\ {}_k\xb^0_t, (1 - \bar{\alpha}_n)\Ib\right)$.\\
        \tcp{prediction}
        ${}_k\hat{\xb}^0_t \gets \hat{\xb}_{\theta_3}({}_k\xb_t^n, n, {}_k\hb_{t_0 : t-1})$\\
        Take gradient step on $\nabla_\theta || {}_k\xb^0_t - {}_k\hat{\xb}^0_t ||^2_2$
    }
\end{algorithm}

\begin{algorithm}[t]
    \caption{Sampling ${}_k\xb^0_t$}
    \label{alg:sample}
    \KwIn{
        Noise: ${}_k\xb^N_t \sim \cN(\zerob, \Ib)$, Context: ${}_k{\yb^0_{t_0:t-1}}$, ${}_k\zb_{t_0:t-1}$, $\sbb_k$, and $\Mb_k$.
    }
    \For{$n \in {N, ..., 1}$}{
        $\ub \gets \zerob$\\
        \If{ $n > 1$}{
            $\ub \sim \cN(\zerob, \Ib)$
        }
        % \Else{
        %     $\ub \gets \zerob$
        % }
        Compute context embedding ${}_k\hb_{t_0 : t-1}$.\\
        ${}_k\hat{\xb}^0_t \gets \hat{\xb}_\theta({}_k\xb_t^n, n, {}_k\hb_{t_0 : t-1})$\\
        ${}_k\xb^{n-1}_t \gets \frac{\sqrt{\alpha_n}(1 - \bar{\alpha}_{n-1})}{1 - \bar{\alpha_n}} {}_k\xb^n_t + \frac{\sqrt{\bar{\alpha}_{n-1}}\beta_n}{1 - \bar{\alpha_n}} {}_k\hat{\xb}^0_t$\\
        $\numquad[16] + \sqrt{\Sigma_\theta}\ub$
    }
    \Return ${}_k\xb^0_t$
\end{algorithm}
To train \dflood{}, we sample the following from the training time-series data of each patch: ($i$) inundation history in the context window (${}_k\xb^0_{t_0 : t-1}$) and temporal co-variates (${}_k\zb_{t_0, t-1}$), ($ii$) inundation values in the immediate next timestep (${}_k\xb^0_{t}$). \dflood{} is trained to forecast the latter.
We apply Algorithm \ref{alg:training} on each sample. First, we generate a random binary sensor mask $\Mb_k$. This mask is applied on the context inundation history (${}_k\xb^0_{t_0 : t-1}$) to generate a sparse inundation history ${}_k\yb^0_{t_0 : t-1}$ (line \ref{algline:data_masking_start}-\ref{algline:data_masking_end}, $\odot$ denotes element-wise multiplication). Here, we retain the inundation values in cells where sensors are placed (according to the sensor mask $\Mb_k$). For cells that do not have sensors, their inundation values are replaced with standard gaussian noise. This serves as a signal to the model to ignore these values. As we train \dflood{} with different random sensor masks, it is able to make forecasts based on different sensor placement configurations during inference.
We use the sparse inundation history, temporal co-variates and the sensor mask to calculate the context embedding ${}_k\hb_{t_0 : t-1}$ (lines \ref{algline:context_embedding_start}-\ref{algline:context_embedding_end}).
We then optimize the model parameters $(\theta_1, \theta_2, \theta_3)$ by minimizing the negative log-likelihood of the data conditioned on the context embedding ($-\log p_\theta({}_k\xb^0_{t} | {}_k\hb_{t_0 : t-1}$). Through a similar derivation as shown in the previous section, the objective to minimize becomes:
\begin{align}
    \EE_{{}_k\xb^0_t, n} || {}_k\xb^0_t - \hat{\xb}_{\theta_3}({}_k\xb_t^n, n, {}_k\hb_{t_0 : t-1} ||^2_2 \label{obj:dflood_obj_1}
\end{align}

Here, the neural network $\hat{x}_{\theta_3}$ is now also conditioned on the context embedding ${}_k\hb_{t_0 : t-1}$. 

\subsection{Inference}
During inference, given a context (${}_k\yb^0_{t_0:t-1}$, ${}_k\zb_{t_0:t-1}$, $\sbb_k$, $\Mb_k$), we first compute the context embedding ${}_k\hb_{t_0 : t-1}$. We then apply Algorithm \ref{alg:sample} to obtain a sample ${}_k\xb^0_t$ of the next time step. Based on this sample, we compute the new context embedding ${}_k\hb_{t_0 : t}$ and repeat until the desired number of prediction time steps have been reached. We repeat the process multiple times to get multiple predictions.
\section{Experiment}
In our experiments, we first evaluate \dflood{} on the TideWatch dataset against six competitive baseline models; taking forecasting inundation within the \textit{Virginia Eastern Shore} region as our case study. Then, we conduct an ablation study to demonstrate the importance of the different context information utilized in \dflood{}. 
% The model's source code is provided in the supplementary material.

\subsection{Dataset}
\label{sec:dataset}
% \zakaria{why this data? why not publicly available dataset?}
% \zakaria{Define the dataset a bit, mention the train-test split}
% source of the data
% details about the data: coverage, number of days, land and sea cells
% How we preprocessed the data - patch definition 30m x 30m resolution, min max normalization
% Train test split
% Elevation data
To train \dflood{}, we downloaded inundation data of the Eastern Shore, Virginia from Tidewatch Maps~\citep{Loftis2019tidewatch,Loftis2022tidewatch}; a street-level inundation mapping tool developed by the Virginia Institute of Marine Science (VIMS). It utilizes the open-source hydro-dynamic model, SCHISM \citep{zhang2016seamless} as the engine, and provides~36 hour inundation forecast maps with a 12 hour update frequency. We used this dataset due to its high spatial resolution, coverage of areas affected by coastal inundation, and lack of other publicly available inundation datasets. Moreover, using it as training data allows \dflood{} to learn to make forecasts from a physics-based hydro-dynamic model. The dataset will be shared upon request.

We downloaded 89 days (March 28, 2024 to June 24, 2024) of hourly inundation data of our study area. The data comes in a raster geo-database format with a pixel resolution of $10m \times 10m$ and provides inundation level at each pixel in feet. We resampled the data to a pixel resolution of $30m \times 30m$. We then consider different square-shaped patches.
% of size: $16 \times 16$, $64 \times 64$, $80 \times 80$, and $96 \times 96$. 
% \zakaria{Visualizations of the patches are provided in the supplementary materials}. 
Each patch $P_k$ has a time series of inundation levels denoted by ${}_k\xb^0_t \in \RR^{D \times D}$, where $D \in \{16,64,80,96\}$ is the side length of the patch. 
% By providing training data for multiple different patches, \dflood{} is trained to make predictions for different patches.
% land and sea cell, we assign max inundation on land cell in training data to sea cells
We also used digital elevation data \citep{DigitalElevation2021} for the same area. We resampled the elevation data to the same resolution and then aligned it with our inundation data. It serves as spatial context (i.e., elevation of patch $P_k$ denoted as $\sbb_k$) within \dflood{}.

We have used 81 days 
% (1944 timesteps, each of length 1-hour) 
of data for training, 1 day 
% (24 timesteps) 
of data for validation, and 7 days 
% (168 timesteps) 
of data for testing. Training data points are sampled by following the process described in Section `\dflood{} Method (Training)'. We use a context window of $12$ hours. During testing, we predict the inundation levels in the subsequent $12$ hours.
% Each sample datapoint therefore contains 24 consecutive timesteps, first $12$ hours for context and the last $12$ hours for prediction. 
Inundation values are standardized based on the mean and standard deviation of the inundation values in the training data. Elevation values are standardized separately based on the mean and standard deviation of the elevation values in the study area.

\begin{table*}[!h]
\centering
\resizebox{\textwidth}{!}{
\begin{tabular}{p{0.06\linewidth}p{0.06\linewidth}p{0.12\linewidth}p{0.12\linewidth} p{0.12\linewidth} p{0.12\linewidth} p{0.12\linewidth}p{0.18\linewidth}|p{0.18\linewidth}}
\toprule
Patch\newline$D^2\times K$ & Metric & DiffSTG & DCRNN & DeepVAR & LSTNet & TimeGrad & BayesNF & {\dflood{}} \\ \toprule

\multirow{2}{*}{$16^2\times2$} & NACRPS & 1.2375\scriptsize{$\pm$} 0.0473 & N/A & 2.9912\scriptsize{$\pm$} 0.09 & N/A & 1.1233\scriptsize{$\pm$} 0.1139 & \textbf{0.714}\scriptsize{$\pm$0.062 ($\uparrow$27.28\%)} & \underline{0.9819}\scriptsize{$\pm$} 0.1248\\

 & NRMSE & \underline{0.1295}\scriptsize{$\pm$} 0.0018 & 0.1643\scriptsize{$\pm$} 0.0003 & 0.2564\scriptsize{$\pm$} 0.0061 & 0.1454\scriptsize{$\pm$} 0 & 0.1504\scriptsize{$\pm$} 0.0082 & \textbf{0.1023}\scriptsize{$\pm$0.006 ($\uparrow$22.38\%)} & 0.1318\scriptsize{$\pm$} 0.0036\\
 %\cmidrule{2-9}

 %& NACRPS &  &  0.0205\scriptsize{$\pm$0.0197} & 0.7063\scriptsize{$\pm$0.0114} & 0.6313\scriptsize{$\pm$0.0297} & \textbf{0.0097\scriptsize{$\pm$0.0016}} & 0.0112\scriptsize{$\pm$0.0006} & 0.0109\scriptsize{$\pm$0.0011} \\  
 
 %& NRMSE &  &  0.0165\scriptsize{$\pm$0.0124} & 0.6027\scriptsize{$\pm$0.0081} & 0.4305\scriptsize{$\pm$0.022} & 0.0129\scriptsize{$\pm$0.0054} & 0.0079\scriptsize{$\pm$0.0004} & \textbf{0.0078}\scriptsize{$\pm$0.0007} \\
 \midrule

\multirow{2}{*}{$16^2\times5$} & NACRPS & 1.2959\scriptsize{$\pm$} 0.0398 & N/A & 2.4144\scriptsize{$\pm$} 0.0462 & N/A & 0.9952\scriptsize{$\pm$} 0.0347 &  \textbf{0.707}\scriptsize{$\pm$} 0.0661 ($\uparrow$24.44\%) & \underline{0.9357}\scriptsize{$\pm$} 0.2\\ 
 
 & NRMSE & 0.1359\scriptsize{$\pm$} 0.0026 & 0.2026\scriptsize{$\pm$} 0.0113 & 0.2329\scriptsize{$\pm$} 0.0042 & 0.1515\scriptsize{$\pm$} 0 & 0.1477\scriptsize{$\pm$} 0.0019 & \textbf{0.1088}\scriptsize{$\pm$} 0.0073 ($\uparrow$13.44\%) & \underline{0.1257}\scriptsize{$\pm$} 0.0157 \\ %\cmidrule{2-9}
 %& NACRPS & 0.0155\scriptsize{$\pm$0.0108} & 0.046\scriptsize{$\pm$0.0403} & 0.6442\scriptsize{$\pm$0.01} & 0.5264\scriptsize{$\pm$0.0034} & 0.0188\scriptsize{$\pm$0.0088} & 0.0107\scriptsize{$\pm$0.0013} & \textbf{0.0093}\scriptsize{$\pm$0.0018} \\  
 %& NRMSE & 0.0166\scriptsize{$\pm$0.0098} & 0.0413\scriptsize{$\pm$0.0344} & 0.5249\scriptsize{$\pm$0.0082} & 0.4143\scriptsize{$\pm$0.0019} & 0.0702\scriptsize{$\pm$0.0294} & 0.0131\scriptsize{$\pm$0.0043} & \textbf{0.0108}\scriptsize{$\pm$0.0031} \\ 
 \midrule

\multirow{2}{*}{$64^2\times10$} & NACRPS & \multirow{2}{*}{Failed} & N/A & 0.967\scriptsize{$\pm$} 0.0124 & N/A & 0.5741\scriptsize{$\pm$} 0.0249 & \underline{0.2668}\scriptsize{$\pm$} 0.0136 & \textbf{0.2028}\scriptsize{$\pm$} 0.0446 ($\uparrow$23.99\%) \\ 
 
 & NRMSE &  & 0.2735\scriptsize{$\pm$} 0.0004 & 0.2907\scriptsize{$\pm$} 0.0021 & 0.2696\scriptsize{$\pm$} 0.0009 & 0.1845\scriptsize{$\pm$} 0.0039 & \underline{0.097}\scriptsize{$\pm$} 0.0047 & \textbf{0.069}\scriptsize{$\pm$} 0.016 ($\uparrow$28.87\%) \\ %\cmidrule{2-9}
 %& NACRPS & 0.0155\scriptsize{$\pm$0.0108} & 0.046\scriptsize{$\pm$0.0403} & 0.6442\scriptsize{$\pm$0.01} & 0.5264\scriptsize{$\pm$0.0034} & 0.0188\scriptsize{$\pm$0.0088} & 0.0107\scriptsize{$\pm$0.0013} & \textbf{0.0093}\scriptsize{$\pm$0.0018} \\  
 %& NRMSE & 0.0166\scriptsize{$\pm$0.0098} & 0.0413\scriptsize{$\pm$0.0344} & 0.5249\scriptsize{$\pm$0.0082} & 0.4143\scriptsize{$\pm$0.0019} & 0.0702\scriptsize{$\pm$0.0294} & 0.0131\scriptsize{$\pm$0.0043} & \textbf{0.0108}\scriptsize{$\pm$0.0031} \\ 
 \midrule
 
\multirow{2}{*}{$64^2\times20$} & NACRPS & \multirow{2}{*}{Failed}  & \multirow{2}{*}{Failed}  & 0.9165\scriptsize{$\pm$} 0.0125 & N/A & {0.6221}\scriptsize{$\pm$} 0.0199 & \underline{0.2826}\scriptsize{$\pm$} 0.0167 &  \textbf{0.2622}\scriptsize{$\pm$} 0.0593 ($\uparrow$7.22\%) \\ 

& NRMSE &  &  & 0.2927\scriptsize{$\pm$} 0.0021 & 0.2735\scriptsize{$\pm$} 0.0009 & {0.2064}\scriptsize{$\pm$} 0.0043 & \underline{0.113}\scriptsize{$\pm$} 0.0057 & \textbf{0.0924}\scriptsize{$\pm$} 0.0192 ($\uparrow$18.23\%) \\ 
%\cmidrule{2-9}
 %&  NACRPS & 0.0233\scriptsize{$\pm$0.017} & 0.0929\scriptsize{$\pm$0.0366} & 0.5897\scriptsize{$\pm$0.0087} & 0.6793\scriptsize{$\pm$0.0039} & 0.025\scriptsize{$\pm$0.0066} & \textbf{0.0109}\scriptsize{$\pm$0.0019} & 0.0114\scriptsize{$\pm$0.0024} \\ 
 %& NRMSE & 0.0207\scriptsize{$\pm$0.0124} & 0.0878\scriptsize{$\pm$0.0163} & 0.4804\scriptsize{$\pm$0.0075} & 0.4235\scriptsize{$\pm$0.0014} & 0.0841\scriptsize{$\pm$0.0192} & \textbf{0.0126}\scriptsize{$\pm$0.0018} & 0.0129\scriptsize{$\pm$0.001} \\ 
 \midrule 
 
\multirow{2}{*}{$80^2\times20$} & NACRPS & \multirow{2}{*}{Failed} & \multirow{2}{*}{Failed} & 0.9028\scriptsize{$\pm$} 0.0123 & N/A & \underline{0.6195}\scriptsize{$\pm$} 0.0271 & \multirow{2}{*}{Failed}  & \textbf{0.2377}\scriptsize{$\pm$} 0.0419 ($\uparrow$61.63\%) \\ 

 & NRMSE &  &  & 0.1994\scriptsize{$\pm$} 0.0013 & 0.1866\scriptsize{$\pm$} 0.0011 & \underline{0.1387}\scriptsize{$\pm$} 0.0035 &  & \textbf{0.0559}\scriptsize{$\pm$} 0.0104 ($\uparrow$59.7\%) \\ 
 %\cmidrule{2-9}
 %& NACRPS & 0.0249\scriptsize{$\pm$0.0137} & 0.0449\scriptsize{$\pm$0.0253} & 0.5823\scriptsize{$\pm$0.0125} & 0.6356\scriptsize{$\pm$0.003} & 0.0361\scriptsize{$\pm$0.0404} & 0.0106\scriptsize{$\pm$0.0012} & \textbf{0.0103}\scriptsize{$\pm$0.0021} \\ 
 %& NRMSE & 0.0239\scriptsize{$\pm$0.0127} & 0.0335\scriptsize{$\pm$0.0195} & 0.476\scriptsize{$\pm$0.009} & 0.4538\scriptsize{$\pm$0.0024} & 0.0956\scriptsize{$\pm$0.0686} & 0.0147\scriptsize{$\pm$0.0037} & \textbf{0.0145}\scriptsize{$\pm$0.0039} \\
 \midrule

 \multirow{2}{*}{$96^2\times20$} & NACRPS & \multirow{2}{*}{Failed} & \multirow{2}{*}{Failed} & 0.9487\scriptsize{$\pm$} 0.0182 & N/A & \underline{0.6641}\scriptsize{$\pm$} 0.0353 & \multirow{2}{*}{Failed} & \textbf{0.2645}\scriptsize{$\pm$} 0.0574 ($\uparrow$60.17\%) \\
 
 & NRMSE &  &  & 0.1091\scriptsize{$\pm$} 0.0008 & 0.103\scriptsize{$\pm$} 0.0002 & \underline{0.0781}\scriptsize{$\pm$} 0.0019 &  & \textbf{0.0312}\scriptsize{$\pm$} 0.0048  ($\uparrow$60.05\%) \\ 
 %\cmidrule{2-9}
 %& NACRPS & 0.0249\scriptsize{$\pm$0.0137} & 0.0449\scriptsize{$\pm$0.0253} & 0.5823\scriptsize{$\pm$0.0125} & 0.6356\scriptsize{$\pm$0.003} & 0.0361\scriptsize{$\pm$0.0404} & 0.0106\scriptsize{$\pm$0.0012} & \textbf{0.0103}\scriptsize{$\pm$0.0021} \\ 
 %& NRMSE & 0.0239\scriptsize{$\pm$0.0127} & 0.0335\scriptsize{$\pm$0.0195} & 0.476\scriptsize{$\pm$0.009} & 0.4538\scriptsize{$\pm$0.0024} & 0.0956\scriptsize{$\pm$0.0686} & 0.0147\scriptsize{$\pm$0.0037} & \textbf{0.0145}\scriptsize{$\pm$0.0039} \\
 \bottomrule

\end{tabular}
}
\caption{Performance Comparison of \dflood{} with baseline models at 95\% sparsity level (i.e., only $5\%$ of cells in a patch have sensors). 
All metrics are reported up to one standard deviation after conducting ten experiments with different random seeds. For each configuration, the best performance is bold-highlighted and the second-best performance is underlined. For \dflood{}, percentage improvement in performance with its closest baseline competitor is shown. State-of-the-art spatiotemporal forecasting methods (e.g., DiffSTG, DCRNN) could not handle training for large patch configurations; these are marked as `Failed'. Since, DCRNN and LSTNet are point forecasting methods, their NACRPS metrics are marked as `N/A'.
}
\label{tab:baseline-results}
\vspace{-0.25cm}
\end{table*}

\subsection{Evaluation Metric} 
%(\zakaria{there was an error that was fixed in the appendix later. Find that. also do not need two separate metric now.})
We use two metrics to evaluate forecast quality. The first metric is Normalized Root Mean Squared Error (NRMSE).
Let $\xb$ be the set of all observations over all cells for all timesteps. Let $x_{ij,t} \in \xb$ denote the observed inundation value of cell $(i, j)$ at time $t$ and let, $\Tilde{x}_{ij,t}$ denote the corresponding average forecast value. Then:
\begin{equation}
    \text{NRMSE} = \tfrac{1}{\max_{ij,t}(\xb) - \min_{ij,t}(\xb)}\sqrt{\tfrac{\sum_{ij, t}{(x_{ij,t}-\Tilde{x}_{ij,t})^2}}{\sum_{ij, t}1}}
\end{equation}
% Here, $|x|$ denotes the total number of cells.
%\zakaria{Use Different notation for number of forecasts}
% \textbf{Note: We inadvertently put the number of forecasts $M$ as denominator inside the root term while defining this metric. It was an oversight on our part and we apologize for that. What we have here is the correct definition.}

While NRMSE captures the mean error between the forecasts and the ground truth, it cannot take into account the uncertainty of the prediction. To capture this, we use an extended version of the Continuous Ranked Probability Score (CRPS)~\cite{winkler1996scoring}. The CRPS between a single observation $x_{ij,t}$ and the empirical CDF $\hat{F}_{ij,t}^M$ of the $M$ corresponding forecasts (detail in supplementary material), denoted as CRPS$(\hat{F}_{ij,t}^M,x_{ij,t})$, is extended for the multivariate case as Normalized Average CRPS (NACRPS).
\begin{equation}
    \text{NACRPS}(\hat{F}^M,\xb) = \tfrac{\sum_{ij,t}{\text{CRPS}(\hat{F}_{ij,t}^M,x_{ij,t})}}{\sum_{ij,t}{|x_{ij,t}|}}
\end{equation}
Here, $|x_{ij,t}|$ denotes the absolute value of $x_{ij,t}$.

\subsection{Baseline Comparison}
We now describe our baseline methods. All our experiments were conducted on an HPC cluster using the SLURM scheduler. Each job used up to $256$GB CPU RAM, $8$ CPU cores and one GPU with maximum $40$ GB GPU RAM. Our source code is provided with the supplementary materials.
\begin{enumerate}[leftmargin=*,noitemsep,nolistsep]
    % \item SFF: Follows the multi-layer perceptron architecture, univariate in nature. In our implementation, we used a two-layered architecture with 256 and 128 hidden dimensions.
    \item DiffSTG~\citep{wen2023diffstg}: The State-of-the-art (SOTA) diffusion-based model for spatiotemporal forecasting.
    \item TimeGrad~\citep{rasul2021autoregressive} The SOTA diffusion-based model designed for multivariate time-series forecasting.
    \item DeepVAR~\citep{salinas2019high}: Combines an RNN-based architecture with a Gaussian copula process output model with a low-rank covariance structure. 
    \item LSTNet~\citep{lai2018modeling}: Captures local dependency pattern among variables and long-term temporal correlation among time-series trends using CNN and RNN.
    \item DCRNN~\citep{li2017diffusion}: A spatiotemporal forecasting framework that combines diffusion convolution and sequence-to-sequence architecture to capture temporal and spatial dependencies.
    \item BayesNF~\citep{saad2024scalable}: SOTA DL method based on a Bayesian Neural Network that maps a multivariate space-time coordinate to a real-valued field. 
    % Rather than being autoregressive by nature, it performs extrapolation in the temporal domain and interpolation across the spatial domain. %Therefore, we do not apply any sensor masking during the evaluation of BayesNF.
\end{enumerate}

%\zakaria{will fix below part later}

% It is worth noting that we also attempted to use $D^3$VAE \cite{li2022generative} as a baseline. However, we found it to scale poorly; it caused Memory Error even for our lowest patch configuration. Therefore, we consider TimeGrad to be the state-of-the-art across the baseline comparison. 

We used six different patch settings in our experiments as shown in Table ~\ref{tab:baseline-results} (visualization of patches provided in supplementary materials). The experiments are performed at a sparsity level of $95\%$; meaning when forecasting for a patch, only $5\%$ of the cells of that patch will have sensors and inundation values in the context window. During training, we generate random sensor masks ($\Mb_k$) that have $\sim5\%$ of the cell values to be $1$. For testing, we generate $10$ different sensor masks at random with same ($95\%$) sparsity level for each patch setting. These masks are applied on test data points (from Tidewatch) in a round-robin manner; i.e., mask 1 applied on data point 1, mask 2 applied on data point 2, ..., mask 10 applied on datapoint 10, mask 1 applied on datapoint 11, and so on. This ensures that all forecasting methods are being tested with same datapoints. By masking the test datapoints and then making forecasts based on them, we simulate the real-world scenario of forecasting based on sparse sensor observations.
% All hyper-parameters of our model are provided in \zakaria{the supplementary material}. 
%In the first setting, we extracted 10 ($K=10$) patches of size $64\times64$ ($D=64$). Then, we increased $K$ to $20$ while keeping the patch size the same as before. Finally, we again extracted 10 patches but increased the patch sizes to $80\times80$.

It should to be noted that the number of variables increases quadratically with patch size and linearly with the number of patches. We found that some models do not scale well\footnote{We considered $D^3$VAE \citep{li2022generative} as a baseline. It scaled poorly; causing memory error for our lowest patch configuration.} with increments in the number of variables. Across all experiments, we set the context and prediction length to 12 (experiment results with varying prediction length are provided in the supplementary materials). Eight scenarios were sampled to understand the quality of probabilistic forecasting. Table~\ref{tab:baseline-results} summarizes the performance of \dflood{} with respect to the baseline methods. Some key observations from the results are as follows:
% \begin{figure}[!b]
%     \centering
%     \includegraphics[width=0.85\linewidth]{Figures/param_scaling_2.png}
%     \caption{Scaling in Model Parameters (Log scale in Y axis) with increasing patch configuration. It has to be noted that model parameter is not affected by $K$. DiffSTG and DCRNN are not shown here since they fail to execute for large patch setting.}
%     \label{fig:scale-param}
% \end{figure}

% \begin{figure}[t]
%     \centering
%     %\hfill
%     \subfloat[Scaling in Model Parameters\label{fig:scale-param}]{
%         \includegraphics[width=0.47\columnwidth]{Figures/param_scaling.png}
%     }
%     \hfill
%     \subfloat[Scaling in Performance\label{fig:scale-perform}]{
%         \includegraphics[width=0.47\columnwidth]{Figures/model_performance_L-NRMSE.png}
%     }
%     \caption{Scaling of our model with the number of variables. The Y-axis is in the Log Scale. The uncertainty region in Figure~\ref{fig:scale-perform} accounts for one standard deviation.}
%     \label{fig:scale}
% \end{figure}

\emph{First,} we observe that DiffSTG, DCRNN and BayesNF scale poorly to large patch configurations. DiffSTG was unable to process patch configurations where $D > 16$. For DCRNN, $(D=64, K=10)$ was the highest patch configuration that was processed successfully, whereas BayesNF fails for $D > 64$. In larger settings, we ran out of GPU memory for these methods. It demonstrates the relatively worse computational scalability of these three methods. 
% \dflood{} and the other three baseline methods (DeepVAR, LSTNet, TimeGrad) were able to successfully process patch configurations upto $(D=96, K=20)$ without any issues. 

\emph{Second,} in small patch configurations, specifically $D=16 \text{ and } K \in \{2, 5\}$, \dflood{} performs second best. Here, BayesNF is the superior model. \dflood{} outperforms the other baselines in these configurations.
% However, in configuration $D=64, K=10$, which is the largest configuration that BayesNF scales to, \dflood{} outperforms BayesNF by a margin of $\sim16\%$ and $\sim26\%$ in terms of NACRPS and NRMSE, respectively.
% Bayesnf does not use any context inundation history. No need to mention auto-regressiveness. This is likely due to the fact that the non-autoregressive nature of BayesNF fails to take the temporal uncertainty associated with extrapolation from past observations. 

\emph{Third,} in moderate and large patch configurations ($D \in \{64, 80, 96\}$, $K \in \{10, 20\}$), \dflood{} outperforms all six baseline methods in terms of both metrics. Across moderate configuration where $D=64$, which is the largest configuration that BayesNF scales to, \dflood{} outperforms BayesNF by a margin of $\sim7$-$24\%$ and $\sim18$-$29\%$ in terms of NACRPS and NRMSE, respectively. In larger configurations ($D \in \{80,96\}, K=20$), \dflood{} achieves $59$-$61\%$ improvement over the second-best method (TimeGrad), in terms of the two performance metrics. This illustrates the superiority of \dflood{} over baseline methods in performance scalability. 
%This highlights the fact that they are highly dependent on the autoregressive correlation of the historical data and fails to capture physical properties from other context in the absence of sensor data.~\dflood{}, 
% on the other hand, is able to leverage physical contexts from elevation data and other temporal covariates, which guides its learning capability into making prediction even under sparse observation setting.  

These experiment results highlight that existing spatiotemporal forecasting methods have either poor computational scalability or performance scalability, making it challenging to apply them to large-scale high-resolution spatiotemporal forecasting tasks. ~\dflood{} offers both computation and performance scalability, making it suitable for large-scale high-resolution spatiotemporal forecasting tasks under sparse observations. 

\begin{figure}[t]
    \centering
    \includegraphics[width=0.85\columnwidth]{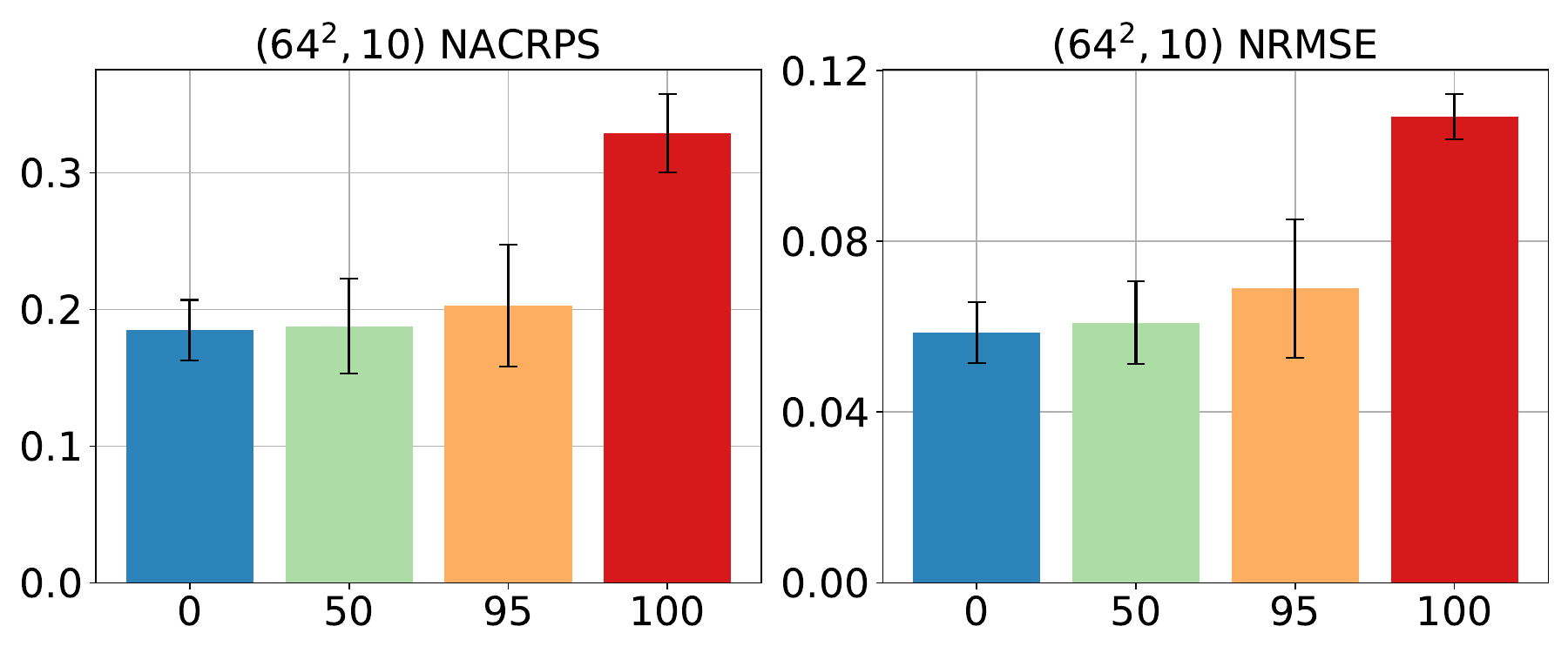}
    \includegraphics[width=0.85\columnwidth]{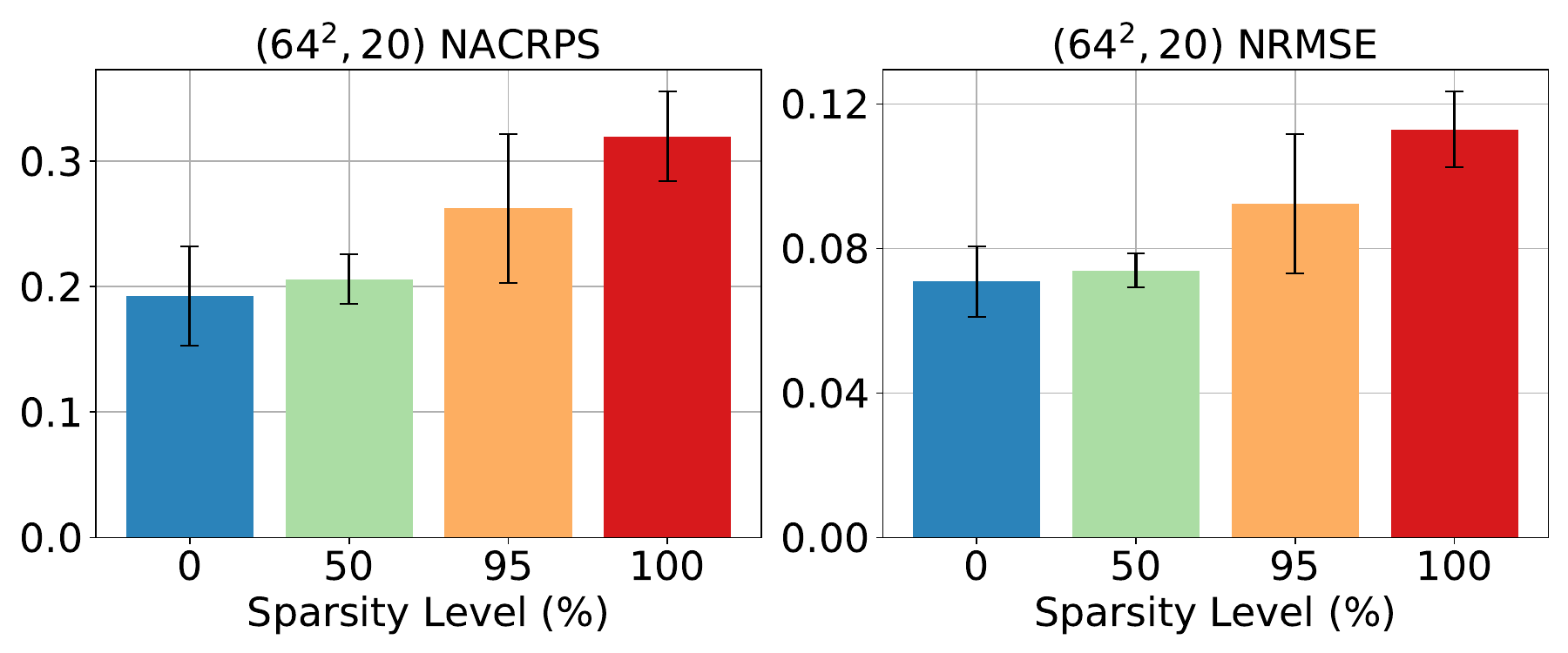}
    \caption{Bar-plots showing \dflood{} performance at varying sparsity levels in two patch configurations. Ten experiment runs were performed in each setting; mean and standard deviation of the performance metrics are shown. 
    % We observe that in the two patch configurations $(64^2, 10)$ and $(64^2, 20)$, as sparsity level increases, performance of \dflood{} degrades (lower is better) in terms of both metrics.
    }
    \label{fig:varying_sparsity_level}
    % \vspace{-0.25cm}
\end{figure}
\subsection{Ablation Study}
In this section, we perform ablation studies on \dflood{} to understand the importance of different context data. First, we investigate the predictive performance of \dflood{} at various sparsity levels. We trained and tested \dflood{} at four different sparsity levels ($0\%$, $50\%$,$95\%$, and $100\%$) for the patch configurations $(64^2, 10)$ and $(64^2, 20)$. Sparsity level of $100\%$ means no sensor data is available. Figure \ref{fig:varying_sparsity_level} shows the performance of \dflood{} in these settings, in terms of our two performance metrics. We observe that in both patch configurations, \dflood{} has the best performance at $0\%$ sparsity level in terms of both metrics. As the sparsity level increases, performance of \dflood{} degrades; the worst performance observed at $100\%$ sparsity level. It implies that, having (more) sensors, i.e., observed inundation values in the context window, helps \dflood{} to make better predictions.

\begin{figure}[t]
    \centering
    \includegraphics[width=0.85\columnwidth]{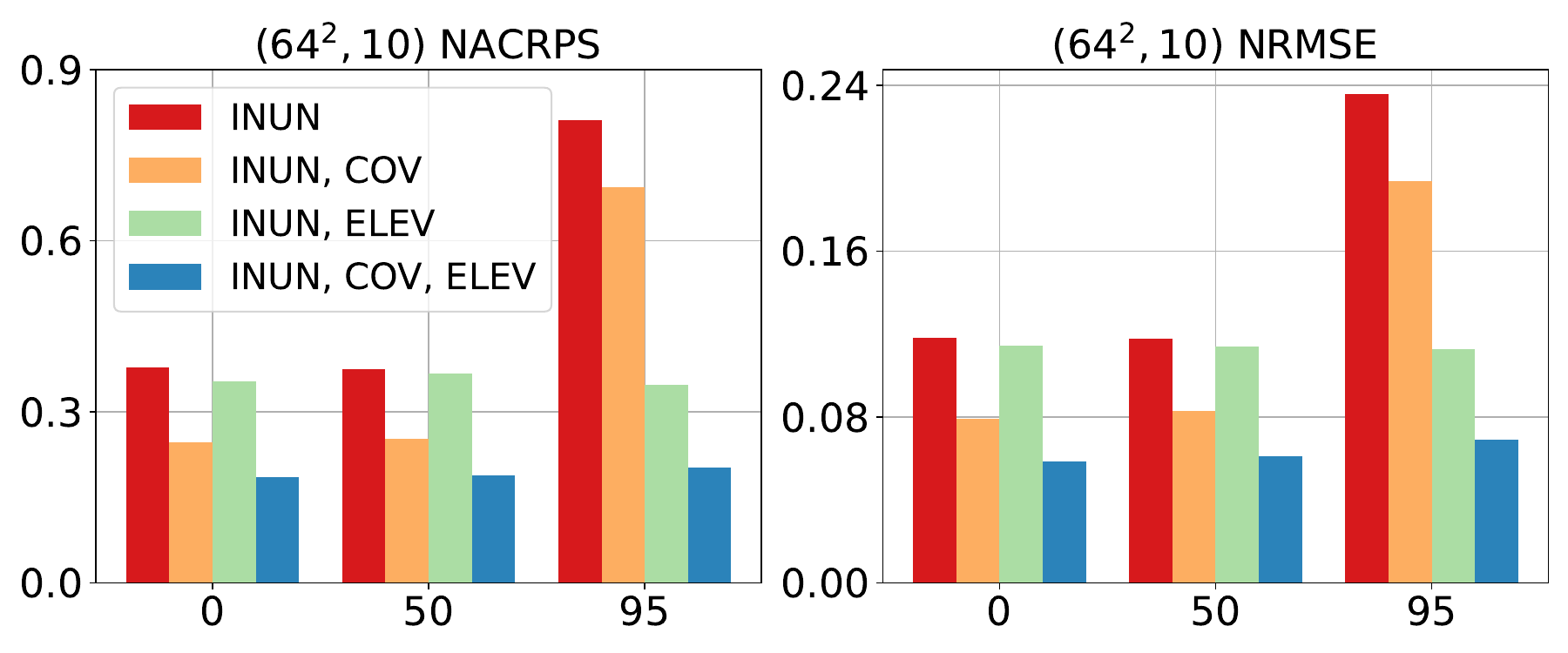}
    \includegraphics[width=0.85\columnwidth]{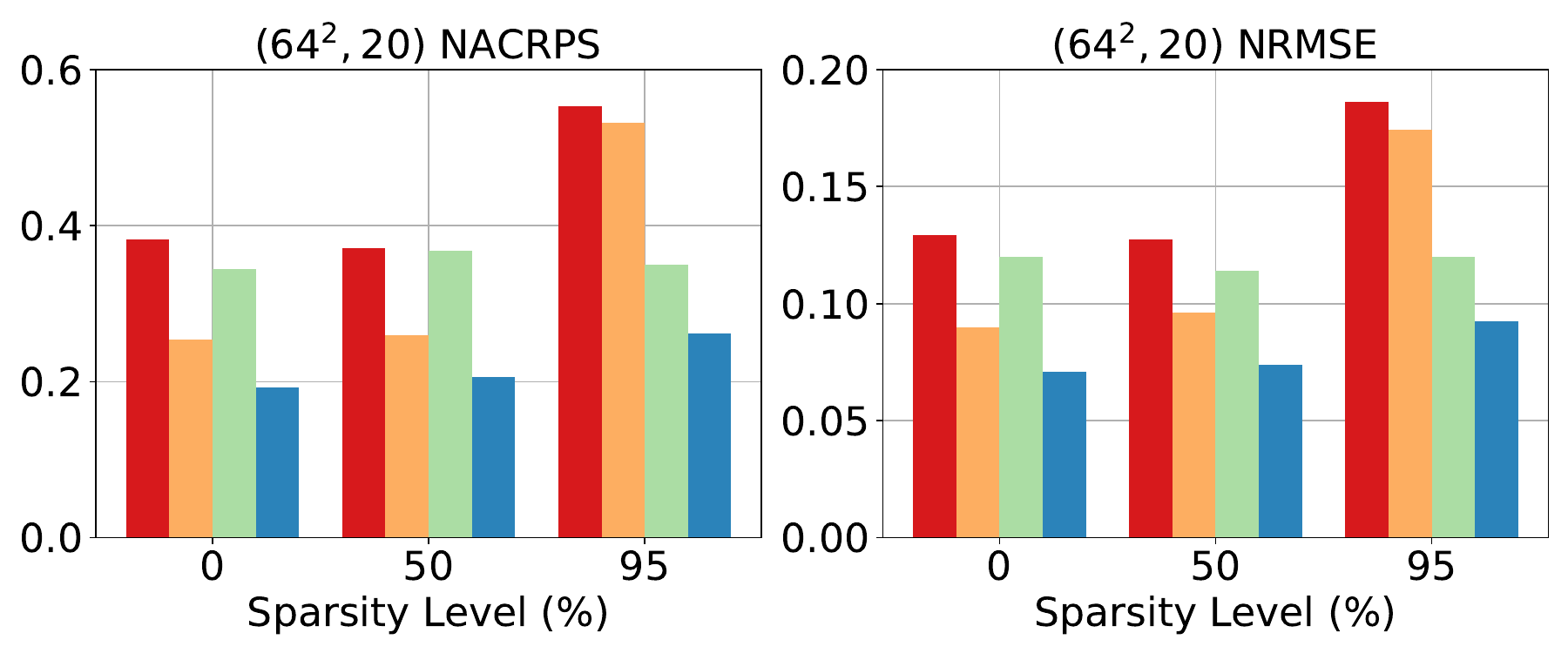}
    \caption{Ablation study at varying sparsity levels for two patch configurations. Ten experiment runs were performed in each setting; mean value of the performance metrics are shown using bar-plots.}
    \label{fig:ablation_at_varying_sparsity_64_10}
    % \vspace{-0.25cm}
\end{figure}
Next, we investigate the utility of elevation data and temporal co-variates as additional context. Given sparse inundation history (INUN: ${}_k\yb^0_{t_0:t-1}$), elevation (ELEV: $\sbb_k$), and temporal co-variates (COV: $\zb_{t_0:t-1}$), we examine four configurations of \dflood{} based on how the context embedding is computed:
(1) INUN: only sparse inundation values used, (2) INUN, ELEV: sparse inundation values and elevation data used, (3) INUN, COV: sparse inundation values and temporal co-variates used, (4) INUN, ELEV, COV: all three components used.
% \begin{enumerate}[leftmargin=*,itemsep=0pt]
%     \item INUN: Only the sparse inundation values used.
%     \item INUN, ELEV: Sparse inundation values and elevation matrix used. 
%     \item INUN, COV: Sparse inundation values and the temporal covariates used.
%     \item INUN, ELEV, COV: All three components used.
% \end{enumerate}
Figure \ref{fig:ablation_at_varying_sparsity_64_10} shows the performance of these four settings at three different sparsity levels ($0\%$, $50\%$, and $95\%$) for the patch configurations $(64^2, 10)$ and $(64^2, 20)$. We see that in all sparsity levels, both (INUN, COV) and (INUN, ELEV) performs better than INUN; meaning both elevation data and temporal co-variates are individually helpful in making better forecasts. At low sparsity levels ($0\%$, $50\%$), the setting (INUN, COV) performs better than (INUN, ELEV), indicating the higher utility of temporal co-variates than elevation at low sparsity levels. However, at high sparsity level ($95\%$), we see the opposite; (INUN, ELEV) performs better than (INUN, COV). It means at high sparsity level, elevation data becomes more useful compared to temporal co-variates. In all sparsity levels, using all three components (i.e., INUN, COV, ELEV) yields the best result. 

\section{Conclusion}
In this paper, we have addressed the problem of high resolution probabilistic coastal inundation forecasting from sparse sensor observations. To solve it, we first formulated it as a problem of learning a conditional distribution function. We then presented \dflood{}, a scalable spatiotemporal forecasting method that utilizes sparse sensor observations, elevation data, and temporal covariates to make probabilistic forecasts. \dflood{} is trained using data generated by a physics-based hydro-dynamical model. A novel masking strategy is employed to tackle the challenge of making forecasts based on sparse sensor observations during inference. Moreover, due to our robust training strategy, \dflood{} can make forecasts based on different sensor placement configurations without requiring retraining from scratch. Through our experiments, we demonstrated that \dflood{} outperforms existing forecasting methods in terms of computational and performance scalability. Our extensive ablation study reveals that all three spatial and temporal context data: elevation data, temporal covariates, and sensor observations (even if sparse), help \dflood{} make better forecasts.

\section{Acknowledgments}
This work was partially supported by the following grants: (i) NSF Grants CCF-1918656, OAC-1916805, RISE-2053013; (ii) AI Institute: Agricultural AI for Transforming Workforce and Decision Support (AgAID) award No. 2021-67021-35344; (iii) U.S. Department of Energy, through the Office of Advanced Scientific Computing Research's “Data-Driven Decision Control for Complex Systems (DnC2S)” project. Pacific Northwest National Laboratory is operated by Battelle Memorial Institute for the U.S. Department of Energy under Contract No. DE-AC05-76RL01830.

\bibliography{aaai2026}

\appendix
% \newpage
\section{Related Work}

%We present an abridged version of the most relevant literature in this section. The supplementary material of this paper covers the literature more extensively.

% \textbf{Hydrology Models} Traditionally, inundation has been simulated through hydrodynamical physics-based 1D/2D dual drainage models. By simulating water flow in water bodies and coupling them with 1D or 2D hydrodynamic models, these models can predict inundation in surfaces. For example, the 2D model SOBEK~\cite{deltares2014sobek} consists of a 1D network with information about the initial water volume of rivers and 2D rectangular computational cells. By coupling these two parts, the model uses the Saint-Venant flow equation and 2D shallow water equation to calculate discharge between cells and determine the water level at each cell of the Digital Elevation Model (DEM). TELEMAC-2D~\cite{vu2015two} also solves 2D shallow water equation to simulate water flow. However, its primary purpose is to understand flow through water bodies. Therefore, this makes this model more suitable for coastal regions, contrary to SOBEK which are more suitable for urban areas. Polymorphic models like SCHISM~\cite{zhang2016seamless} can capture water flow at very high resolution, down to even a few meters due to their hybrid grinding system and localized coordinate system. Despite these models' ability to model inundation accurately at fine resolution, the computation time associated with them make them infeasible for real-time forecasting.

\noindent
\textbf{Hydrology Models:} %Traditionally, various Physics Based hydrodynamic Models (PBM) have been employed to simulate inundation. SOBEK~\citep{deltares2014sobek} is a 2D model that solves a Saint-Venant flow equation and shallow water equation to determine the water level of rectangular grids. TELEMAC-2D~\citep{vu2015two} uses a similar method but is more suitable for coastal areas. Polymorphic models like SCHISM~\citep{zhang2016seamless} can capture water flow at very high resolution. However, the computation time associated with solving these complex equations at high spatial resolution makes these PBMs infeasible for real-time forecasting. 
Traditionally, inundation has been simulated through hydro-dynamical physics-based 1D/2D dual drainage models. By simulating water flow in water bodies and coupling them with 1D or 2D hydrodynamic models, these models can predict inundation in surfaces. For example, the 2D model SOBEK~\citep{deltares2014sobek} consists of a 1D network with information about the initial water volume of rivers and 2D rectangular computational cells. By coupling these two parts, the model uses the Saint-Venant flow equation and 2D shallow water equation to calculate discharge between cells and determine the water level at each cell of the Digital Elevation Model (DEM). TELEMAC-2D~\citep{vu2015two} also solves 2D shallow water equation to simulate water flow. However, its primary purpose is to understand flow through water bodies. Therefore, this makes this model more suitable for coastal regions, contrary to SOBEK which is more suitable for urban areas. Polymorphic models like SCHISM~\citep{zhang2016seamless} can capture water flow at very high resolution, down to even a few meters due to their hybrid grinding system and localized coordinate system. Despite these models' ability to model inundation accurately at fine resolution, the computation time associated with them makes them infeasible for real-time forecasting ~\citep{guo2021urban}.

\noindent
\textbf{Regression-based Models}
Early approaches in finding surrogates for hydrology models involved applying various regression-based methods. Chang~\emph{et al.}~\citep{chang2010clustering} first applied K-means clustering to pre-process data to find inundation control points. The authors then applied linear regression or neural networks to predict inundation for grids depending on whether they are linearly correlated with the control points or not. Bermúdez~\emph{et al.}~\citep{bermudez2019rapid} explored non-parametric regression approach based on least square support vector machine (SVM) as surrogate for 2D hydrodynamic models. Lee~\emph{et al.}~\citep{lee2021scenario} created an offline rainfall-runoff-inundation database by simulating hydrology models over multiple scenarios. Then, they trained a logistic regression-based approach based on the database to estimate the parameters to predict the risk of flooding for each region. Their goal was then to use the trained regressor to generate near real-time prediction of inundation probability given the runoff to the grid from simulated and realistic rainfall scenario.

\noindent
\textbf{Deep Learning:} DL methods are a popular choice in surrogating hydrodynamic models. Pan~\emph{et al.}~\citep{pan2011hybrid} applied a combined approach consisting of principal component analysis (PCA) and a feed-forward neural network. Their approach involved inundation prediction at various locations in the next timestep by looking at a fixed window of rainfall history of $N$ different rainfall gauges. First, they applied PCA on these various rainfall gauge histories to reduce the dimensionality before using them as input to the feed-forward neural network. Since such neural networks are expected to fail where data availability is limited, Xie~\emph{et al.}~\citep{xie2021artificial} explored the potential of a hybrid feed-forward neural network-based approach where they used two block-based neural networks along with the traditional point-based neural network and found superiority of the hybrid approach over other in the data-sparse region. Consequently, convolutional neural network (CNN) was explored~\citep{kabir2020deep} as a potential surrogate. After their model was trained to predict the output simulated from a 2D hydrology model, they used it to simulate two real-event flood scenarios in a region of UK. Since these models are not able to capture long-term temporal dependencies, researchers have recently explored~\citep{roy2023application} seq2seq LSTM model as surrogates to hydrology models. Such model was also applied in forecasting for multi-step future horizons. Aside from this, regression-based methods~\citep{zahura2022predicting} and hybrid ML-based methods~\citep{zanchetta2022hybrid} have been attempted as possible surrogates. However, these models cannot inherently take spatial context into account. The DCRNN method proposed by~\citep{li2017diffusion} for traffic forecasting is the state-of-the-art DL method for spatiotemporal point forecasting, where bidirectional random walk is used to capture spatial dependency and Seq2Seq architecture is used to capture temporal dependency. 

%Many DL models have been explored as possible surrogates for Physics Based Models (PBM). 

\noindent
\textbf{Diffusion Model} The field of generative modeling has been dominated by the diffusion model for quite some time. Although it had primarily been studied for image synthesis~\citep{ho2020denoising,dhariwal2021diffusion,austin2021structured}, its fascinating ability to use flexible conditioning to guide the generation process has been utilized in other domains like video generation~\citep{yang2023diffusion}, prompt-based image generation~\citep{ramesh2022hierarchical}, text-to-speech generation~\citep{yang2023diffsound}. TimeGrad~\citep{rasul2021autoregressive} was the pioneering work that proposed the use of diffusion models for time-series forecasting tasks. They used temporal co-variates and embedded historical data as conditions for the diffusion model to denoise the prediction for the next time step. CSDI~\cite{tashiro2021csdi} does the same but instead of encoding with a Recurrent Neural Network (RNN) as done in TimeGrad, they leverage a transformer-based architecture. SSSD$^{S4}$~\cite{alcaraz2022diffusion} uses an S4 model~\cite{gu2021efficiently} to capture the encoding which is particularly useful in capturing long-term dependencies. Also, instead of performing diffusion along both temporal and input channel dimensions, they perform diffusion only along the temporal dimension.  Other models like DPSD-CSPD~\citep{bilovs2023modeling}, TDSTF~\citep{Chang2024} were later introduced. These models try to predict the noise added during the forward process of the diffusion model ($\epsilon$-parameterization). An alternative parameterization is to predict the ground-truth data ($x$-parameterization). 
% We utilize the $x$-parameterization to accommodate domain-specific constraints in the training objective. 
Different from models mentioned above, $D^3$VAE~\citep{li2022generative} uses coupled diffusion process and bidirectional variational autoencoder to better forecast in longer horizon and improve interpretability. The State-of-the-Art Diffusion-based model DiffSTG~\citep{wen2023diffstg} devised a modified architecture for the reverse process of diffusion model combining UNet and GNN to capture temporal and spatial dependencies, respectively. However, both these methods scale poorly with the growth in number of variables. BayesNF~\citep{saad2024scalable} is the state-of-the-art DL spatiotemporal forecasting model, which is built using a Bayesian Neural Network. By assigning a prior distribution to the model parameters and adjusting the posterior based on observation, BayesNF essentially learns to map multivariate spatiotemporal points to a continuous real-valued field. 
\section{Denoising Diffusion Model}
\label{sec:diffusion_basic_app}
In this section, we provide an overview of denoising diffusion models.

Let $\xb^0 \sim q_\chi(\xb^0)$ denote the multivariate training vector from some input space $\chi = \mathbb{R}^D$ and true distribution $q_\chi(\xb^0)$. In diffusion probabilistic models, we aim to approximate $q_\chi(\xb^0)$ by a probability density function $p_{\theta}(\xb^0)$ where $\theta$ denotes the set of trainable parameters. Diffusion probabilistic models \cite{sohl2015deep, luo2022understanding} are a special class of Hierarchical Variational Autoencoders \cite{kingma2016improved, sonderby2016ladder} with the form $p_{\theta}(\xb^0) = \int p_\theta(\xb^{0:N})d\xb^{1:N}$; here $\xb^1, \xb^2, ..., \xb^N$ are latent variables. Three key properties of diffusion models are: $(i)$ all of the latent variables  are assumed to have the dimension $D$ as $\xb^0$, $(ii)$ the approximate posterior $q(\xb^{1:N}|\xb^0)$, $$\textstyle q(\xb^{1:N}|\xb^0) = \prod_{n=1}^N q(\xb^n | \xb^{n-1})$$ is fixed to a Markov chain (often referred to as the \textit{forward} process), and $(iii)$ the structure of the latent encoder at each hierarchical step is pre-defined as a linear Gaussian model as follows: $$q(\xb^n|\xb^{n-1}) = \cN(\xb^n; \sqrt{1-\beta_n}\xb^{n-1}, \beta_n\Ib).$$ Additionally, the Gaussian parameters ($\beta_1, \beta_2, ..., \beta_n)$ are chosen in such a way that we have: $q(\xb^N) = \cN(\xb^N; \zerob,\Ib)$. 

The joint distribution $p_{\theta}(\xb^{0:N})$ is called the reverse process. It is defined as a Markov chain with learned Gaussian transitions beginning at $p(\xb^N)$. $$\textstyle p_\theta(\xb^{0:N}) = p(\xb^N) \prod_{n=1}^N p_{\theta}(\xb^{n-1}|\xb^n).$$
Each subsequent transition is parameterized as follows:  
\smallskip
\begin{align}
    p_\theta(\xb^{n-1}|\xb^n) = \cN(\xb^{n-1}; \mu_\theta(\xb^n, n), \Sigma_\theta(\xb^n, n)\Ib) \label{eq:backward_gaussian_supp}   
\end{align}
Both $\mu_\theta: \RR^D \times \NN \rightarrow \RR^D$ and $\Sigma_\theta: \RR^D \times \NN \rightarrow \RR^+$ take two inputs: $\xb^n \in \RR^D$ and the noise step $n \in \NN$. The parameters $\theta$ are learned by fitting the model to the data distribution $q_\chi(\xb^0)$. This is done by minimizing the negative log-likelihood via a variational bound:
\begin{align}
    \log p_\theta(\xb^0) &= \log \int p_\theta(\xb^{0:N})d\xb^{1:N} \nonumber\\
    & = \log \EE_{q(\xb^{1:N}|\xb^0)} \frac{p_\theta(\xb^{0:N})}{q(\xb^{1:N}|\xb^0)} \nonumber\\
    & \geq \EE_{q(\xb^{1:N}|\xb^0)} \log\frac{p_\theta(\xb^{0:N})}{q(\xb^{1:N}|\xb^0)} \label{eq:jenesen_apply_supp}\\
    \implies -\log p_\theta(\xb^0) &\leq \EE_{q(\xb^{1:N}|\xb^0)} \log\frac{q(\xb^{1:N}|\xb^0)}{p_\theta(\xb^{0:N})} \label{eq:variational_bound_supp}
    % & \quad\EE_{q(\xb^{1:N}|\xb^0)} \left[-\log p_\theta(\xb^{0:N}) + \log q(\xb^{1:N}|\xb^0)\right] 
\end{align}

Here, we get (\ref{eq:jenesen_apply_supp}) by applying Jensen's inequality. We indirectly minimize the negative log-likelihood of the data, i.e., $-\log p_\theta(\xb^0)$, by minimizing the right hand side of (\ref{eq:variational_bound_supp}) . We can show that it is equal to:
\begin{equation}
    \EE_{q(\xb^{1:N}|\xb^0)} \left[ -\log p_\theta(\xb^N) + \sum_{n=1}^N \log \frac{q(\xb^n|\xb^{n-1})}{p_\theta(\xb^{n-1}|\xb^n)} \right] \label{eq:obj_1_supp}
\end{equation}

\cite{ho2020denoising} showed that the forward process has the following property: we can sample $\xb^n$ using $\xb^0$ at any noise step $n$ in closed form . Let, $\alpha_n = 1 - \beta_n$, and $\bar{\alpha}_n = \prod_{i=1}^n \alpha_i$. Then, we have:
\begin{equation}
    q(\xb^n | \xb^0) = \cN\left(\xb^n; \sqrt{\bar{\alpha}_n}\xb^0, (1 - \bar{\alpha}_n)\Ib\right). \label{eq:forward_property_supp}
\end{equation}

The authors then showed that (\ref{eq:obj_1_supp}) can be written as KL-divergence between Gaussian distributions:
\begin{align}
    & -\EE_{q(\xb^1|\xb^0)}\log p_\theta(\xb^0|\xb^1) + D_{KL}\left( q(\xb^N|\xb^0)\ ||\ p_\theta(\xb^N) \right) \nonumber\\
    & + \sum_{n=2}^N \EE_{q(\xb^n|\xb^0)}\left[ D_{KL} (q(\xb^{n-1}|\xb^n, \xb^0)\ ||\ p_\theta(\xb^{n-1}|\xb^n)) \right] \label{eq:obj_2_supp}
\end{align}

% Details below can be moved to supplementary materials
Here, the authors conditioned the forward process posterior on $\xb^0$, i.e. $q(\xb^{n-1}|\xb^n, \xb^0)$, because then it becomes tractable as follows:
\begin{align}
    q(\xb^{n-1}|\xb^n, \xb^0) = \cN(\xb^{n-1}; \tilde{\mu}_n(\xb^n, \xb^0), \tilde{\beta}_n\Ib) \label{eq:forward_posterior_supp}
\end{align}

where, 
\begin{align}
    \tilde{\mu}_n(\xb^n, \xb^0) &= \frac{\sqrt{\alpha_n}(1 - \bar{\alpha}_{n-1})}{1 - \bar{\alpha_n}} \xb^n +
    \frac{\sqrt{\bar{\alpha}_{n-1}}(1-\alpha_n)}{1 - \bar{\alpha_n}} \xb^0 \label{eq:mu_formula_supp}\\
    \tilde{\beta}_n &= \frac{1 - \bar{\alpha}_{n-1}}{1 - \bar{\alpha_n}} \beta_n
\end{align}

Since, the forward process posterior (\ref{eq:forward_posterior_supp}) and the backward transition (\ref{eq:backward_gaussian_supp}) are both Gaussian, we can write the KL-divergence between them as follows:

\begin{align}
    & D_{KL}\left( q(\xb^{n-1}|\xb^n, \xb^0)\ ||\ p_\theta(\xb^{n-1}|\xb^n) \right) \nonumber\\
    & = \frac{1}{2\tilde{\beta}_n} || \tilde{\mu}_n(\xb^n, \xb^0) - \mu_{\theta}(\xb^n, n)||_2^2 + C \label{eq:mu_parameterization_supp}
\end{align}

Here, $C$ is a constant independent of $\theta$, and we assume that 
\begin{align}
    \Sigma_\theta=\tilde{\beta_n} = \frac{1 - \bar{\alpha}_{n-1}}{1 - \bar{\alpha_n}} \beta_n \label{eq:sigma_theta}
\end{align} 
As $\mu_\theta(\xb^n, n)$ is conditioned on $\xb^n$, similar to $\tilde{\mu}_n(\xb^n, \xb^0)$ in (\ref{eq:mu_formula_supp}), we can give it the form:
\begin{align}
    &\mu_\theta(\xb^n, n) = \frac{\sqrt{\alpha_n}(1 - \bar{\alpha}_{n-1})}{1 - \bar{\alpha_n}} \xb^n + \frac{\sqrt{\bar{\alpha}_{n-1}}\ \beta_n}{1 - \bar{\alpha_n}} \hat{\xb}_\theta(\xb^n, n) \label{eq:x_parameterization_supp}
    % &\quad\quad\frac{\sqrt{\bar{\alpha}_{n-1}}(1-\alpha_n)}{1 - \bar{\alpha_n}} \hat{\xb}_\theta(\xb^n, n) \label{eq:x_parameterization}
\end{align}

Here, $\hat{\xb}_\theta(\xb^n, n)$ is a neural network that predicts $\xb^0$ from noisy vector $\xb^n$ and noise step $n$. With this parameterization, we can show that:
\begin{align}
    & D_{KL}\left( q(\xb^{n-1}|\xb^n, \xb^0)\ ||\ p_\theta(\xb^{n-1}|\xb^n) \right) \nonumber\\
    & = \frac{1}{2\tilde{\beta}_n}\frac{\bar{\alpha}_{n-1}\ \beta_n^2}{(1 - \bar{\alpha}_n)^2}|| \xb^0 - \hat{\xb}_\theta(\xb^n, n) ||_2^2
    % &\quad\quad\frac{1}{2\tilde{\beta}_n}\frac{\bar{\alpha}_{n-1}(1-\alpha_n)^2}{(1 - \bar{\alpha}_t)^2}|| \xb^0 - \hat{\xb}_\theta(\xb^n, n) ||_2^2
\end{align}

Therefore, optimizing a diffusion model can be done  by training a neural network to predict the original ground truth vector from an arbitrarily noisy version of it. To approximately minimize the summation term in the objective (\ref{eq:obj_2_supp}) across all noise steps $n$, we can minimize the expectation over all noise steps as follows:
\begin{align}
    \argmin_\theta \ \EE_{n \sim U[2, N]}\  \EE_{q(\xb^n|\xb^0)}\frac{1}{2\tilde{\beta}_n}\frac{\bar{\alpha}_{n-1}\beta_n^2}{(1 - \bar{\alpha}_n)^2}|| \xb^0 - \hat{\xb}_\theta(\xb^n, n) ||_2^2
    % &\quad\quad\frac{1}{2\tilde{\beta}_n}\frac{\bar{\alpha}_{n-1}\ (1-\alpha_n)^2}{(1 - \bar{\alpha}_n)^2}|| \xb^0 - \hat{\xb}_\theta(\xb^n, n) ||_2^2
    % &\quad\quad D_{KL}( q(\xb^{n-1}|\xb^n, \xb^0)\ || \ p_\theta(\xb^{n-1}|\xb^n))
\end{align}
This can be done by using stochastic samples over noise steps. Once trained, to sample from the reverse process $\xb^{n-1} \sim p_\theta(\xb^{n-1}|\xb^n)$, we first sample $\xb^N$ from $\cN(\zerob, \Ib)$. Then, we compute:
\begin{align}
    \xb^{n-1} = \mu_\theta(\xb^n, n) + \sqrt{\Sigma_\theta}\ \ub \label{eq:backward_sample_supp}
\end{align}
Here, $\ub \sim \cN(\zerob, \Ib)$ for $n \in [2, N]$ and $\ub=\zerob$ when $n=1$. $\mu_\theta(\xb^n, n)$ and $\Sigma_\theta$ are computed using (\ref{eq:x_parameterization_supp}) and (\ref{eq:sigma_theta}), respectively. 
% $\xb^0$ is taken as the generated sample.

\section{Evaluation Metric} 
%(\zakaria{there was an error that was fixed in the appendix later. Find that. also do not need two separate metric now.})
We use two metrics to evaluate forecast quality. The first metric is Normalized Root Mean Squared Error (NRMSE).
Let $\xb$ be the set of all observations over all cells for all timesteps. Let $x_{ij,t} \in \xb$ denote the observed inundation value of cell $(i, j)$ at time $t$ and let, $\Tilde{x}_{ij,t}$ denote the corresponding average forecast value. Then:
\begin{equation}
    \text{NRMSE} = \frac{1}{\max_{ij,t}(\xb) - \min_{ij,t}(\xb)}\sqrt{\frac{\sum_{ij, t}{(x_{ij,t}-\Tilde{x}_{ij,t})^2}}{\sum_{ij, t}1}}
\end{equation}
% Here, $|x|$ denotes the total number of cells.
%\zakaria{Use Different notation for number of forecasts}
% \textbf{Note: We inadvertently put the number of forecasts $M$ as denominator inside the root term while defining this metric. It was an oversight on our part and we apologize for that. What we have here is the correct definition.}

 In addition to NRMSE, we use an extended version of the Continuous Ranked Probability Score (CRPS)~\cite{winkler1996scoring} to quantify forecast uncertainty. 
 %can capture the mean error between the forecast and the ground truth, it cannot take into account the uncertainty of the prediction. To capture this, we  
 The CRPS metric is calculated for each cell and each timestep individually. Let's assume that we have $M$ forecasts ${\hat{x}_{ij,t}^1,\hat{x}_{ij,t}^2,...,\hat{x}_{ij,t}^M}$ for observation $x_{ij,t}$. \emph{First,} the empirical cumulative distribution function (CDF) at a point $z$, $\hat{F}_{ij,t}^M(z)$ from these forecast values is defined as:
\begin{equation}
    \hat{F}_{ij,t}^M(z) = \frac{1}{M}\sum_{m=1}^N{\mathbb{I}(\hat{x}_{ij,t}^n \leq z)}
\end{equation}
\noindent
where $\mathbb{I}(y)$ is the indicator function that yields $1$ if condition $y$ holds, $0$ otherwise. From this, CRPS between the empirical CDF $\hat{F}_{ij,t}^M$ and the observation $x_{ij,t}$ is calculated as
\begin{equation}
    \text{CRPS}(\hat{F}_{ij,t}^M,x_{ij,t}) = \int_z{\Big(\hat{F}_{ij,t}^M(z) - \mathbb{I}({x}_{ij,t} \leq z)\Big)^2 dz}
\end{equation}

Since CRPS is defined for individual timestep and individual cells, we extend this metric to Normalized Average CRPS (NACRPS), defined as
\begin{equation}
    \text{NACRPS}(\hat{F}^M,\xb) = \frac{\sum_{ij,t}{\text{CRPS}(\hat{F}_{ij,t}^M,x_{ij,t})}}{\sum_{ij,t}{|x_{ij,t}|}}
\end{equation}
Here, $|x_{ij,t}|$ denotes the absolute value of $x_{ij,t}$.

% \newpage
\section{Additional Experiment Results}
\subsection{Varying Prediction Length}
\begin{figure}[!b]
    \centering
    \includegraphics[width=\columnwidth]{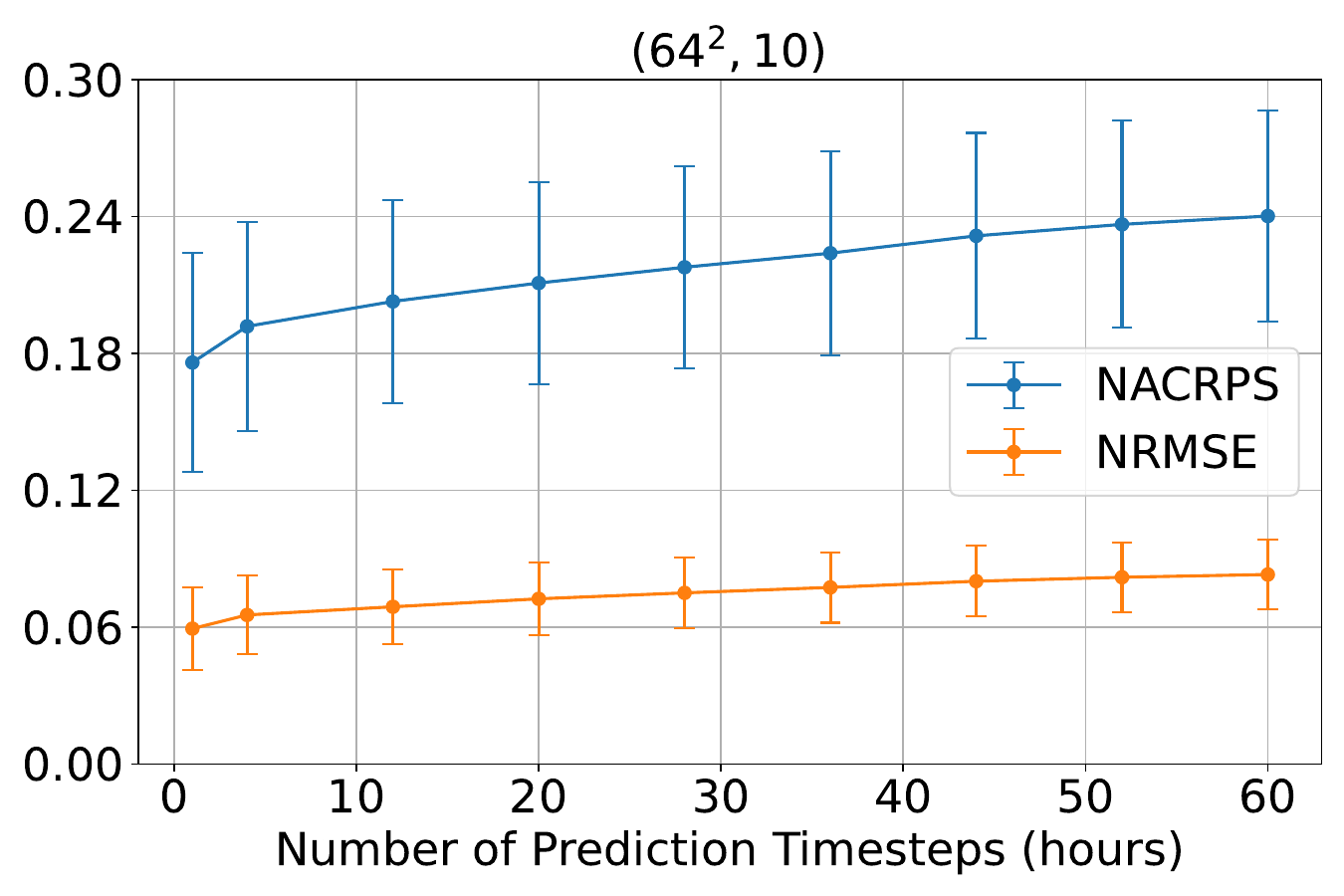}
    \caption{Line-plots showing the performance of \dflood{} for different prediction lengths with test dataset configuration $(64^2, 10)$, context length of 12 hours, and sparsity level of $95\%$, in terms of two metrics. The dots represent the mean value of the metrics over ten experiment runs with different seeds; the vertical  lines represent standard deviation. We observe that \dflood{} performs best when prediction length is $1$. This is expected as \dflood{} is trained to make predictions for one future timestep. When making predictions for multiple timesteps auto-regressively, we observe that both NACRPS and NRMSE increase with increasing number of prediction timesteps.}
    \label{fig:varying_prediction_length_64_10}
\end{figure}

As \dflood{} forecasts the inundation levels in future timesteps auto-regressively, we can use it to forecast inundation levels for arbitrary number of future timesteps. However, this can result in error accumulation. To test this, we took the setting $(64^2, 10)$ and varied the prediction length with the values $1, 4, 12, 20, 28, 36, 44, 52, 60$. Figure \ref{fig:varying_prediction_length_64_10} shows the values of the two performance metrics (NACRPS and NRMSE) for our test data, for each of these prediction lengths, where context length is set to 12 hours and sparsity level to $95\%$. We observe that \dflood{} performs best, in terms of both performance metrics, when prediction length is $1$. This is expected as \dflood{} is trained to make predictions for one future timestep. As the prediction length is increased, we observe that both NACRPS and NRMSE increases. 

\subsection{Masking Strategy}
In \dflood{}, we have presented a novel masking strategy (Algorithm \ref{alg:training}, lines \ref{algline:data_masking_start}-\ref{algline:data_masking_end}). In this strategy, we retain the inundation values in cells where sensors are placed (according to the sensor mask). For cells that do not have sensors, their inundation values are replaced with standard gaussian noise. We refer to this masking strategy as \textit{"Noise Masking"}. To show the efficacy of this masking strategy, we consider a baseline masking strategy as follows: retain the inundation values in cells where sensors are placed; cells that do not have sensors, their inundation values are replaced with the value $0$. We refer to this baseline masking strategy as \textit{"Zero Masking"}.
\begin{figure}[t]
    \centering
    \includegraphics[width=\columnwidth]{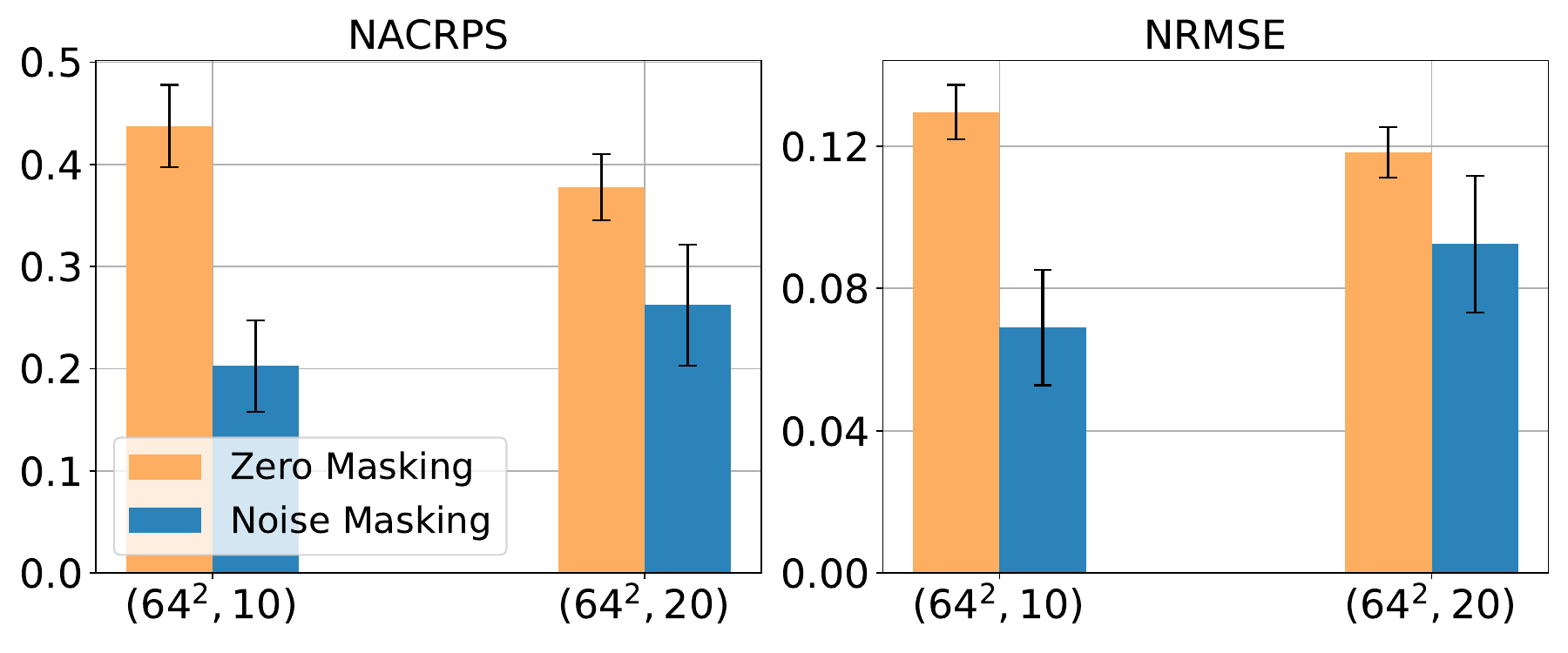}
    \caption{Bar-plots showing the performance of \dflood{} (in terms of NACRPS and NRMSE) for two different masking strategies during training (Zero mask and Noise mask), with patch configuration $(64^2, 10)$ and $(64^2, 20)$ and sparsity level $95\%$. Context and prediction lengths are set to 12 hours. The height of the bars represent the mean value of the metrics over ten experiment runs with different seeds; the vertical black lines represent standard deviation. We observe that \dflood{} noise masking yields better result in terms of both performance metrics for the two patch configurations.}
    \label{fig:varying_masking_strategy}
\end{figure}
Figure \ref{fig:varying_masking_strategy} shows a comparison of the two masking strategies in terms of the resulting predictive performance of \dflood{}. We observe that, for the patch configurations $(64^2, 10)$ and $(64^2, 20)$ at $95\%$ sparsity level, \textit{noise masking} yields better predictive performance than \textit{zero masking}. Our hypothesis is, by placing random gaussian noise, \textit{noise masking} provides an additional signal to the model to ignore the inundation values where sensors are not available.

\subsection{Forecasting Time}
To understand the efficacy of \dflood{} in real-time forecasting, we recorded its forecasting time. With context length of 12 hours, prediction length of 36 hours, and number of scenarios to be predicted set to $8$; \dflood{} takes $\sim 7$ seconds to forecast for a patch of size $64 \times 64$, on a single GPU. Since each cell of the patch is $30m \times 30m$, area of a $64 \times 64$ patch is $1.423$ sq. miles. The entire eastern shore of Virginia has a total area of $2105$ sq miles, which would require $\sim 1480$ patches of size $64 \times 64$. On a single GPU, making 36 hour forecasts for all these patches would take $2.88$ hours using \dflood{}. However, we can parallelize the forecasting of different patches. For instance, if we have $8$ GPUs available, making $36$ hour forecast for the entire Eastern Shore of Virginia would take $2.87 / 8 \approx 0.36$ hours or $\sim 22$ minutes. 

\section{Inundation Scenario Queries}
Once \dflood{} is trained, we can use it to answer inundation scenario-based queries relevant to policymakers. In this Section, we discuss a few of such queries.

% \madhav{Ashik: once completed let me know. I will share the paper with folks. I will add a few more things as well if I find time.}

\begin{figure}[t]
    \centering
    \includegraphics[width=0.492\columnwidth]{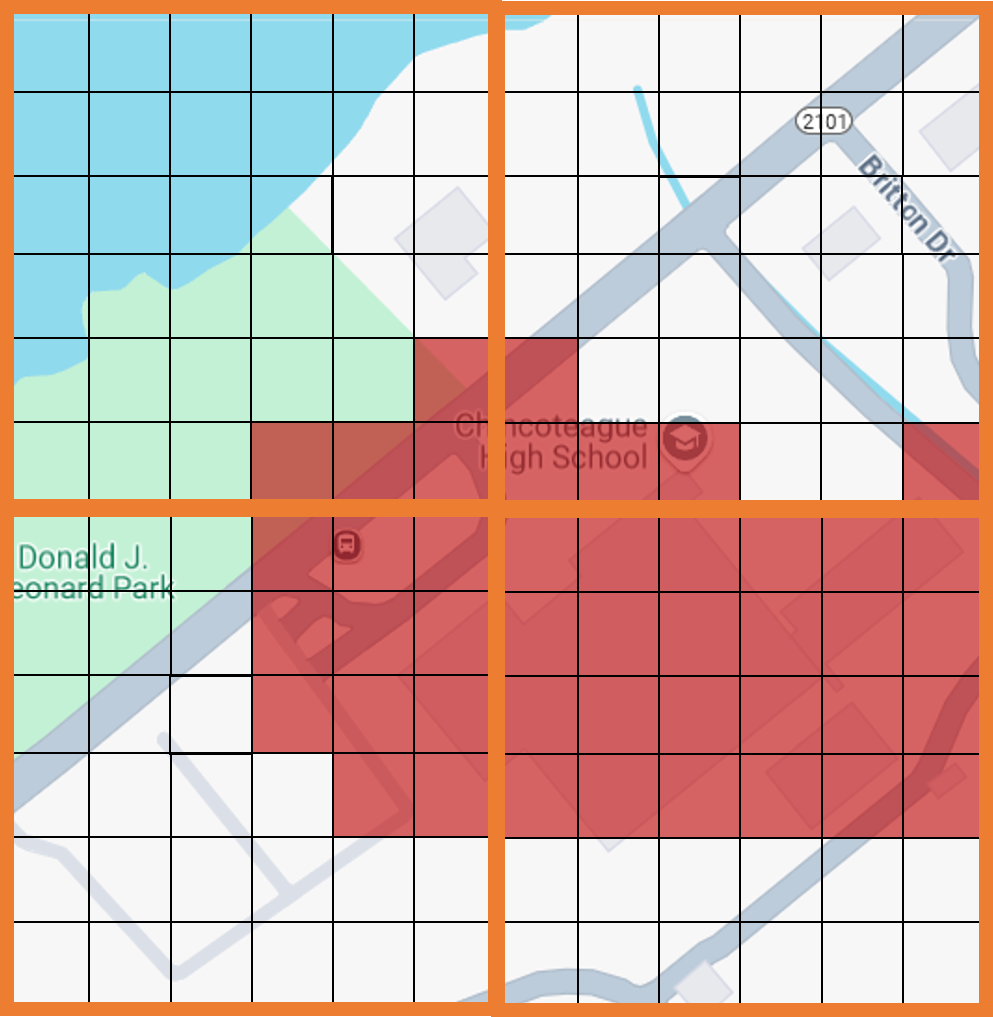}
    \hfill
    % \hspace{0.05\linewidth}
    \includegraphics[width=0.492\columnwidth]{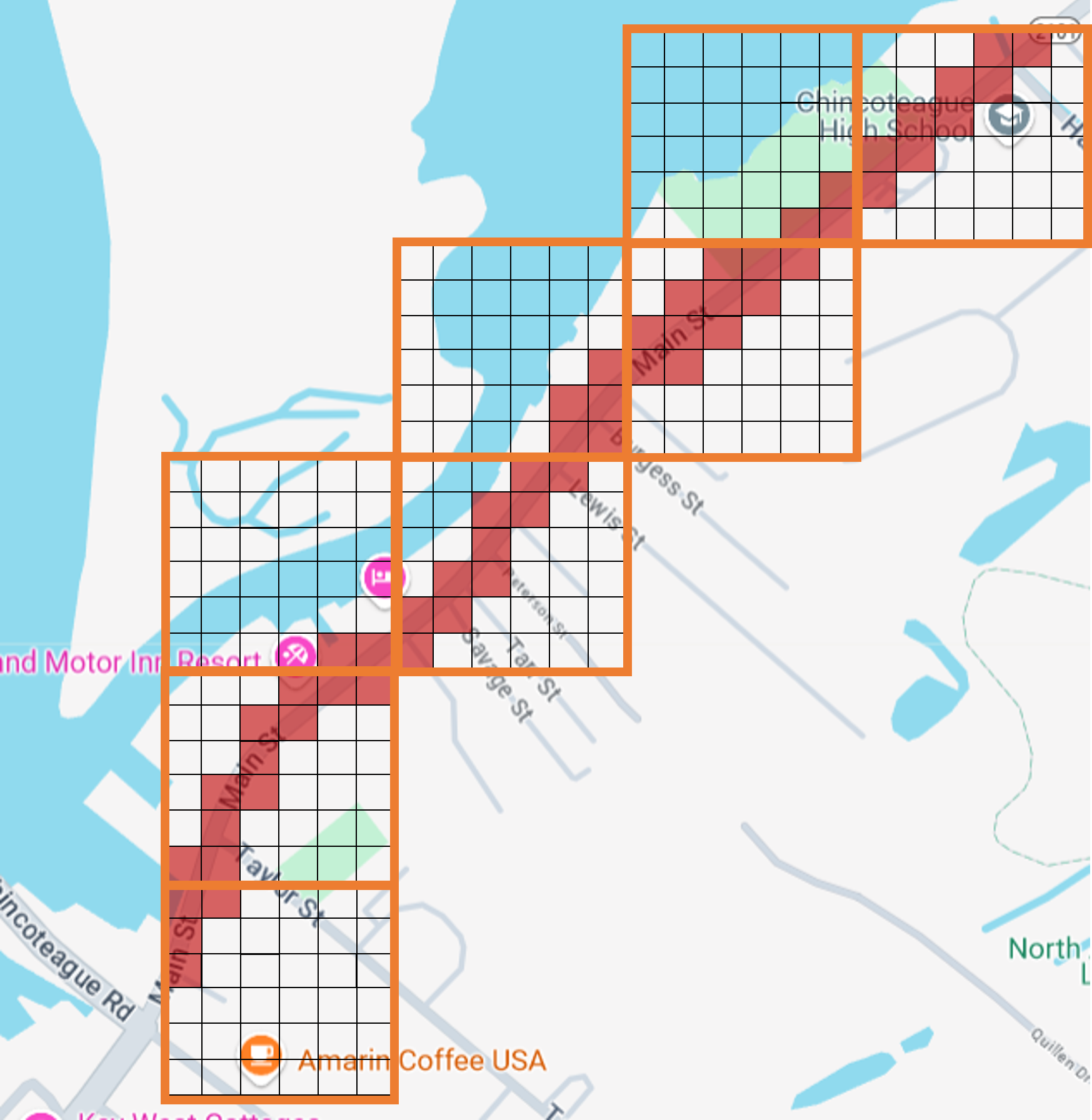}
    \caption{\textbf{(left)} A school area in the Eastern Shore of Virginia. The red cells and four orange patches overlap with the school area. A possible query is: what is the probability that the flooding level in the school area will be above $d$ units within next $T$ hours? \textbf{(right)} A route coming out from the school. The red cells and the eight orange patches overlap with the route. A possible query is: what is the probability that the route coming out of the school will not be flooded in the next $T$ hours?}
    \label{fig:flooding_query_supp}
\end{figure}

\begin{algorithm}[!b]
    \caption{Calculate the probability that the flooding level in an area $A$ will be $\leq d$ units in the next $T$ hours.}
    \label{alg:region_flooding_probability}
    \KwIn{
        Query area $A$ as a Polygon,
        Flooding level threshold $d$,
        Prediction Length $T$,
        Trained \dflood{} Model.
    }
    $C_A \gets$ Set of grid cells that overlap with $A$. \label{algline:identify_cells}\\
    $\cP_A \gets$ Set of patches where each patch $P \in \cP_A$ contains at-least one cell $c \in C_A$. \label{algline:identify_patches}\\
    \For{each patch $P \in \cP_A$}{
        Generate $M$ samples for patch $P$ with prediction length $T$ (using Algorithm 2 in main paper).\label{algline:generate_samples}\\
        $num\_scenarios(P_{\leq d}) \gets$ Number of samples where each cell $c \in P \cap C_A$ has flooding level $\leq d$. \label{algline:calc_num_scenarios}\\
        $probability(P_{\leq d}) \gets num\_scenarios(P_{\leq d}) / M$ \label{algline:calc_patch_prob}
    }
    $p \gets \prod_{P \in \cP_A} probability(P_{\leq d})$ \label{algline:calc_final_prob}\\
    \Return $p$.
\end{algorithm}

\begin{itemize}[leftmargin=*]
\item \textbf{Query 1:} \textit{What is the probability that the flooding level in an area $A$ will be above $d$ units within next $T$ hours?}\\
We use Algorithm \ref{alg:region_flooding_probability} as a sub-routine to answer this query. First, we identify the cells that overlap with $A$ (red cells in Figure \ref{fig:flooding_query_supp} \textbf{left}). Let $C_A$ denote the set of these cells. If at-least one cell $c \in C_A$ has a flooding level above $d$ units in the next $T$ hours, we say area $A$ has a flooding level above $d$ units.

% \madhav{you can define this a bit differently where in a certain number of cells have to be flooded. There is also the issue of temporal consistency when you consider multiple cells. These are called Threhshold queries and as you have done below}
% \kai{Requiring multiple cells to be flooded will also complicate the probability calculation. Let's assume, for example, the cells are spread over 2 patches ($P_1, P_2$) and we require $N$ cells to be flooded to consider the region to be flooded. Then flooding can happen if $(0,N)$ or $(1, N-1) or (2, N-2) or ... (N,0)$ cells are flooded in $P_1, P_2$ respectively. I think we can keep the current query for simplicity and mention that we can have more complex queries.}

In Algorithm \ref{alg:region_flooding_probability} line \ref{algline:identify_patches}, we identify the patches that contain at-least one cell $c \in C_A$ (four orange patches in Figure \ref{fig:flooding_query_supp}, \textbf{left}). Let the set of such patches be denoted by $\cP_A$. For each patch $P \in \cP_A$, we generate $M$ number of samples (using Algorithm 2 in main paper) with prediction length of $T$ hours (line \ref{algline:generate_samples}). Then, we calculate the number of samples where all cells $c \in P \cap C_A$ have a inundation level $\leq d$ units (line \ref{algline:calc_num_scenarios}). We denote it by $num\_scenarios(P_{\leq d})$. Subsequently, we calculate the probability that all the cells $c \in P \cap C_A$ have flooding level $\leq d$ units to be $probability(P_{\leq d}) = num\_scenarios(P_{\leq d}) / M$ (line \ref{algline:calc_patch_prob}). We assume the inundation levels in different patches to be independent. Therefore, the probability that all cells $c \in C_A$ have flooding level $\leq d$ units is calculated to be $p = \prod_{P \in \cP_A} probability(P_{\leq d})$ (line \ref{algline:calc_final_prob}). 

Since we require the probability of flooding level being \textbf{above} $d$ units in area $A$, our desired probability value is $(1-p)$.

\item \textbf{Query 2:} \textit{Given an evacuation route, what is the probability that the route will not be flooded in the next $T$ hours?}\\
An evacuation route is a sequence of roads that are used for evacuating out of an area. We represent each road using a polygon. Then, a route is also a polygon, which is the union of all of its member road polygons. Let the route polygon be denoted by $A$. Then, we use Algorithm \ref{alg:region_flooding_probability} with $d=0$, to calculate our desired probability value. 
\end{itemize}

It is important to note that, we can have more complex queries. For instance, a more refined version of Query 1 can be: \textit{what is the probability that at-least half of the region $A$ will have flooding level above $d$ units in the next $T$ hours?}  The key idea for answering such queries is that, we have a sampling method in the form a trained \dflood{} model and we can use it to sample inundation scenarios from the distribution of possible scenarios. We use the sampled scenarios to calculate our desired probability value.

\section{Reproducibility Checklist Additional Information}
Our source code for \dflood{} and comparison with baseline methods are provided with the supplementary materials. The TideWatch inundation dataset will be provided upon request.
% \newpage
\section{Hyper-parameters}
\begin{table}[!h]
    \centering
    \begin{tabular}{p{5.3cm}r}
        \toprule 
        \textbf{Hyper-parameter} & \textbf{Value} \\
        \toprule
        Learning rate & 0.001\\
        \midrule
        Learning rate update policy & Reduce on Plateau.\\
        \midrule
        Learning rate update patience & 3 epochs.\\
        \midrule
        Learning rate update factor & 0.5\\
        \midrule
        Number of epochs & 40\\
        \midrule
        Training batch size & 32\\
        \midrule
        Validation and test batch size & 4\\
        \midrule
        Context length & 12 timesteps\\
        \midrule
        Training prediction length & 1 timestep\\
        \midrule
        Validation and test prediction length & 12 timesteps\\ 
        \midrule
        Number of diffusion steps & 20\\
        \midrule
        Diffusion $\beta_{min}$ & $10^{-4}$\\
        \midrule
        Diffusion $\beta_{max}$ & 1\\
        \midrule
        \# of scenarios generated for validation & 2\\
        \midrule 
        \# of scenarios generated for test & 8\\
        \bottomrule
    \end{tabular}
    \caption{Shared hyper-parameters for all of our patch configurations in Table 1 of main paper.}
    \label{tab:shared_hp}
\end{table}
\begin{table*}[t]
    \centering
    \begin{tabular}{p{4.5cm}ccccc}
    \toprule
    \textbf{Hyper-parameter} & $D = 16$ & $D=64$ & $D=80$ & $D=96$ \\ \toprule
    Number of Convolution Blocks & 1 & 3 & 3 & 3 \\ \midrule 
    Convolution block number of Channels & [64] & [16, 32, 64] & [16, 32, 64] & [16, 32, 64] \\ \midrule
    Context embedding dimension & 32 & 32 & 64 & 96 \\ \midrule
    UNet Number of Down \newline (and Up) blocks & 4 & 4 & 4 & 4 \\ \midrule
    Number of ResNet layers \newline per UNet block & 2 & 2 & 2 & 2 \\ \midrule
    UNet Down Blocks: \newline Number of Channels & [16, 32, 32, 64] & [16, 32, 32, 64] & [16, 32, 32, 64] & [16, 32, 32, 64] \\ \midrule
    UNet Number of Cross-\newline attention Down / Up Blocks & 2 & 2 & 2 & 2 \\ \midrule
    Group Normalization: \newline Number of Groups & 8 & 8 & 8 & 8 \\
    \bottomrule
    \end{tabular}
    \caption{Hyper-parameters for different patch configurations}
    \label{tab:my_label}
\end{table*}

\newpage
\section{Patch Visualization}
\begin{figure*}[!b]
    \centering
    \includegraphics[width=\textwidth]{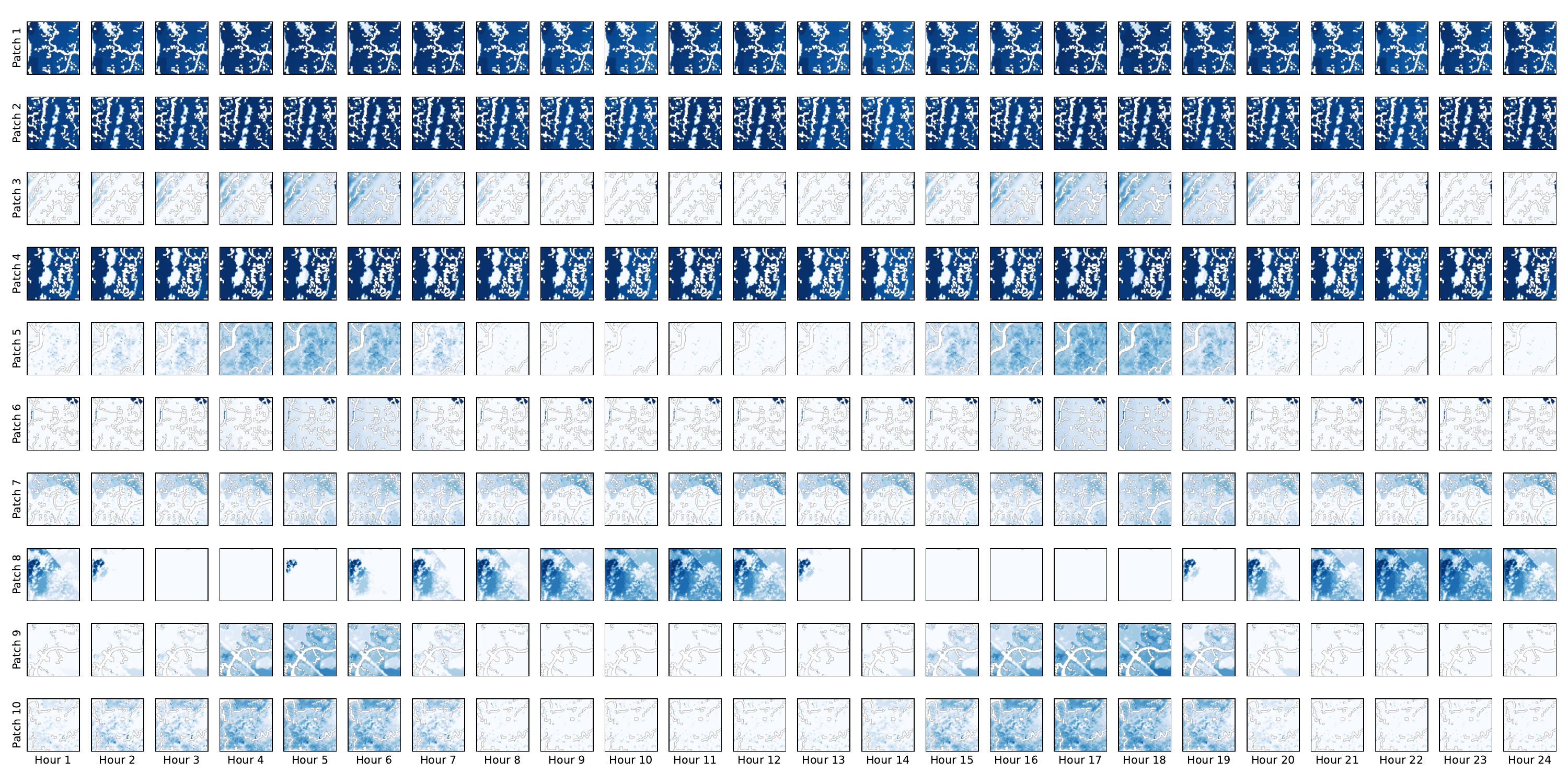}
    \caption{Visualization of inundation on $10$, $64 \times 64$ patches in the first $24$ hours of training data. The darker the shade of blue, the higher the inundation value.}
    \label{fig:patch_vis}
\end{figure*}

\end{document}